\documentclass[runningheads]{llncs}
\usepackage{graphicx}
\usepackage{comment}
\usepackage{amsmath,amssymb} % define this before the line numbering.
\usepackage{color}
\usepackage{booktabs}
\usepackage{algorithm}
\usepackage{algpseudocode}

\usepackage[export]{adjustbox} % Vertical alignment
\usepackage{array,multirow}

\newcommand{\myparagraph}[1]{\vspace{0pt}\noindent{\bf #1}}

\begin{document}
% \linenumbers
\pagestyle{headings}
\mainmatter
\def\ECCVSubNumber{4304}  % Insert your submission number here

\title{GANwriting: Content-Conditioned Generation of Styled Handwritten Word Images}

% \title{GANwriting: Style and Textual Content Conditioned Generation of Handwritten Word Images}
%\title{GANwriting: Content and Style Aware Generation of Handwritten Words}
%\title{(Random) Unsupervised Zero-Shot Multi-Writer Adversarial Handwritten Text Image Generation}

\begin{comment}
\titlerunning{ECCV-20 submission ID \ECCVSubNumber} 
\authorrunning{ECCV-20 submission ID \ECCVSubNumber} 
\author{Anonymous ECCV submission}
\institute{Paper ID \ECCVSubNumber}
\end{comment}
%******************


\titlerunning{GANwriting: Content-Conditioned Generation of Styled Handwriting}
\author{Lei Kang$^{* \dag}$, Pau Riba$^{*}$, Yaxing Wang$^{*}$, Mar\c{c}al Rusi{\~n}ol$^{*}$, Alicia Forn\'{e}s$^{*}$, Mauricio Villegas$^{\dag}$}
% $^{*}$Computer Vision Center, Universitat Aut{\`o}noma de Barcelona, Spain\\
% {\tt\small \{lkang, priba, yaxing, marcal, afornes\}@cvc.uab.es}
% \\
% $^{\dag}$omni:us, Berlin, Germany\\
% {\tt\small \{lei, mauricio\}@omnius.com}
% }

\institute{$^{*}$Computer Vision Center, Universitat Aut{\`o}noma de Barcelona, Spain\\
{\tt\small \{lkang, priba, yaxing, marcal, afornes\}@cvc.uab.es}
\\
$^{\dag}$omni:us, Berlin, Germany\\
{\tt\small \{lei, mauricio\}@omnius.com}\\
}

\authorrunning{L. Kang et al.}

\begin{comment}
\author{First Author\inst{1}\orcidID{0000-1111-2222-3333} \and
Second Author\inst{2,3}\orcidID{1111-2222-3333-4444} \and
Third Author\inst{3}\orcidID{2222--3333-4444-5555}}
%

\institute{Princeton University, Princeton NJ 08544, USA \and
Springer Heidelberg, Tiergartenstr. 17, 69121 Heidelberg, Germany
\email{lncs@springer.com}\\
\url{http://www.springer.com/gp/computer-science/lncs} \and
ABC Institute, Rupert-Karls-University Heidelberg, Heidelberg, Germany\\
\email{\{abc,lncs\}@uni-heidelberg.de}}
\end{comment}
%******************
\maketitle
\begin{abstract}
Although current image generation methods have reached impressive quality levels, they are still unable to produce plausible yet diverse images of handwritten words. On the contrary, when writing by hand, a great variability is observed across different writers, and even when analyzing words scribbled by the same individual, involuntary variations are conspicuous. In this work, we take a step closer to producing realistic and varied artificially rendered handwriting. We propose a novel method that is able to produce credible handwritten word images by conditioning the generative process with both calligraphic style features and textual content. Our generator is guided by three complementary learning objectives: to produce realistic images, to imitate a certain handwriting style and to convey a specific textual content. Our model is unconstrained to any predefined vocabulary, being able to render whatever input word. Given a sample writer, it is also able to mimic its calligraphic features in a few-shot setup. We significantly advance over prior art and demonstrate with qualitative, quantitative and human-based evaluations the realistic aspect of our synthetically produced images.
\end{abstract}

%%%%%%%%%%%%%%%%%%%%%%%%%%%%%%%%%%%%%%%%%%%%%%%%%%%%%%%%%%%%%%%%%%%%%%%%%
%%%%%%%%%%%%%%%%%%%%%%%%%%%%%%%%%%%%%%%%%%%%%%%%%%%%%%%%%%%%%%%%%%%%%%%%%
\section{Introduction}
Few years after the conception of Generative Adversarial Networks (GANs)~\cite{goodfellow2014generative}, we have witnessed an impressive progress on generating illusory plausible images. From the early low-resolution and hazy results, the quality of the artificially generated images has been notably enhanced. We are now able to synthetically produce high-resolution~\cite{brock2018large} artificial images that are indiscernible from real ones to the human observer~\cite{karras2019style}. 
In the original GAN architecture, inputs were randomly sampled from a latent space, so that it was hard to control which kind of images were being generated. The conception of conditional Generative Adversarial Networks (cGANs)~\cite{mirza2014conditional} led to an important improvement. By allowing to condition the generative process on an input class label, the networks were then able to produce images from different given types~\cite{choi2018stargan}. However, such classes had to be predefined beforehand during the training stage and thus, it was impossible to produce images from other unseen classes during inference.

\begingroup
\setlength{\tabcolsep}{8pt} % Default value: 6pt
\begin{figure}[t!]
    \centering
    \begin{tabular}{ccccc}
    \includegraphics[width=0.15\linewidth,height=.8cm]{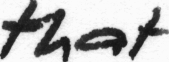}&
    \includegraphics[width=0.15\linewidth,height=.8cm]{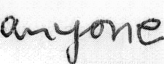}&
    \includegraphics[width=0.15\linewidth,height=.8cm]{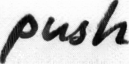}&
    \includegraphics[width=0.15\linewidth,height=.8cm]{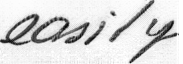}&
    \includegraphics[width=0.15\linewidth,height=.8cm]{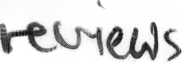}\\
    \includegraphics[width=0.15\linewidth,height=.8cm]{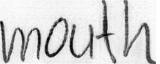}&
    \includegraphics[width=0.15\linewidth,height=.8cm]{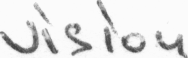}&
    \includegraphics[width=0.15\linewidth,height=.8cm]{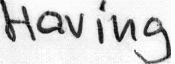}&
    \includegraphics[width=0.15\linewidth,height=.8cm]{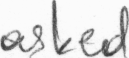}&
    \includegraphics[width=0.15\linewidth,height=.8cm]{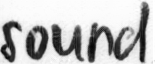}\\
    \includegraphics[width=0.15\linewidth,height=.8cm]{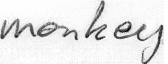}&
    \includegraphics[width=0.15\linewidth,height=.8cm]{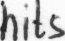}&
    \includegraphics[width=0.15\linewidth,height=.8cm]{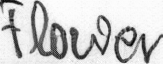}&
    \includegraphics[width=0.15\linewidth,height=.8cm]{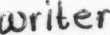}&
    \includegraphics[width=0.15\linewidth,height=.8cm]{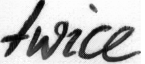}\\
    \end{tabular}
    \caption{Turing's test. Just five of the above words are real. Try to distinguish them from the artificially generated samples\protect\footnotemark[1].}
    \label{fig:turinggame}
\end{figure}
\endgroup

But generative networks have not exclusively been used to produce synthetic images. The generation of data that is sequential in nature has also been largely explored in the literature. Generative methods have been proposed to produce audio signals~\cite{dong2018musegan}, natural language excerpts~\cite{yu2017seqgan}, video streams~\cite{tulyakov2018mocogan} or stroke sequences~\cite{graves2013generating,ha2017neural,ganin2018synthesizing,zheng2019strokenet} able to trace sketches, drawings or handwritten text. In all of those approaches, in order to generate sequential data, the use of Recurrent Neural Networks (RNNs) has been adopted.

Yet, for the specific case of generating handwritten text\footnotetext[1]{\rotatebox[origin=c]{180}{The real words are: \texttt{"that"}, \texttt{"vision"}, \texttt{"asked"}, \texttt{"hits"} and \texttt{"writer"}.}}, one could also envisage the option of directly producing the final images instead of generating the stroke sequences needed to pencil a particular word. Such non-recurrent approach presents several benefits. First, the training procedure is more efficient since recurrencies are avoided and the inherent parallelism nature of convolutional networks is leveraged. Second, since the output is generated all at once, we avoid the difficulties of learning long-range dependencies as well as vanishing gradient problems. Finally, online training data (pen-tip location sequences), which is hard to obtain, is no longer needed. 

Nevertheless, the different attempts to directly generate raw word images present an important drawback. Similarly to the case with cGANs, most of the proposed approaches are just able to condition the word image generation to a predefined set of words, limiting its practical use. For example~\cite{gregor2015draw} is specifically designed to generate isolated digits, while~\cite{chang2018generating} is restricted to a handful of Chinese characters. To our best knowledge, the only exception to that is the approach by Alonso \emph{et al.}~\cite{alonso2019adversarial}. In their work they propose a non-recurrent generative architecture conditioned to input content strings. By having such design, the generative process is not restricted to a particular predefined vocabulary, and could potentially generate any word. However, the produced results are not realistic, still exhibiting a poor quality, sometimes producing barely legible word images. Their proposed approach also suffers from the mode collapse problem, tending to produce images with a unique writing style. In this paper we present a non-recurrent generative architecture conditioned to textual content sequences, that is specially tailored to produce realistic handwritten word images, indistinguishable to humans. Real and generated images are actually difficult to tell apart, as shown in Fig.~\ref{fig:turinggame}. In order to produce diverse styled word images, we propose to condition the generative process not only with textual content, but also with a specific writing style, defined by a latent set of calligraphic attributes.

Therefore, our approach\footnote{Our code is available at \url{https://github.com/omni-us/research-GANwriting}} is able to artificially render realistic handwritten word images that match a certain textual content and that mimic some style features (text skew, slant, roundness, stroke width, ligatures, etc.) from an exemplar writer. To this end, we guide the learning process by three different learning objectives~\cite{odena2017conditional}. First, an adversarial discriminator ensures that the images are realistic and that its visual appearance is as closest as possible to real handwritten word images. Second, a style classifier guarantees that the provided calligraphic attributes, characterizing a particular handwriting style, are properly transferred to the generated word instances. Finally, a state-of-the-art sequence-to-sequence handwritten word recognizer~\cite{michael2019evaluating} controls that the textual contents have been properly conveyed during the image generation. 
To summarize, the main contributions of the paper are the following:
\begin{itemize}
    \item Our model conditions the handwritten word generative process both with calligraphic style features and textual content, producing varied samples indistinguishable by humans, surpassing the quality of the current state-of-the-art approaches.
    \item We introduce the use of three complementary learning objectives to guide different aspects of the generative process.
    \item We propose a character-based content conditioning that allows to generate any word, without being restricted to a specific vocabulary. 
    \item We put forward a few-shot calligraphic style conditioning to avoid the mode collapse problem.
\end{itemize}

%%%%%%%%%%%%%%%%%%%%%%%%%%%%%%%%%%%%%%%%%%%%%%%%%%%%%%%%%%%%%%%%%%%%%%%%%
%%%%%%%%%%%%%%%%%%%%%%%%%%%%%%%%%%%%%%%%%%%%%%%%%%%%%%%%%%%%%%%%%%%%%%%%%
\section{Related Work}
The generation of realistic synthetic handwritten word images is a challenging task. To this day, the most convincing approaches involved an expensive manual intervention aimed at clipping individual characters or glyphs~\cite{wang2005combining,konidaris2007keyword,lin2007style,thomas2009synthetic,haines2016my}. When such approaches were combined with appropriate rendering techniques including ligatures among strokes, textures and background blending, the obtained results were indeed impressive. Haines \emph{et al.}~\cite{haines2016my} illustrated how such approaches could artificially generate indistinguishable manuscript excerpts as if they were written by Sir Arthur Conan Doyle, Abraham Lincoln or Frida Kahlo. Of course such manual intervention is extremely expensive, and in order to produce large volumes of manufactured images the use of truetype electronic fonts has also been explored~\cite{krishnan2016generating,kang2020unsupervised}. Although such approaches benefit from a greater scalability, the realism of the generated images clearly deteriorates.

With the advent of deep learning, the generation of handwritten text was approached differently. As shown in the seminal work by Alex Graves~\cite{graves2013generating}, given a reasonable amount of training data, an RNN could learn meaningful latent spaces that encode realistic writing styles and their variations, and then generate stroke sequences that trace a certain text string. However, such sequential approaches~\cite{graves2013generating,ha2017neural,ganin2018synthesizing,zheng2019strokenet} need temporal data, obtained by recording with a digital stylus pen real handwritten samples, stroke-by-stroke, in vector form.

Contrary to sequential approaches, non-recurrent generative methods have been proposed to directly produce images. Both variational auto-encoders~\cite{kingma2013auto} and GANs~\cite{goodfellow2014generative} were able to learn the MNIST manifold and generate artificial handwritten digit images in the original publications. With the emergence of cGANs~\cite{mirza2014conditional}, able to condition the generative process on an input image rather than a random noise vector, the adversarial-guided image-to-image translation problem started to rise. Image-to-image translation has since been applied to many different style transfer applications, as demonstrated in~\cite{isola2017image} with the \emph{pix2pix} network. Since then, image translation approaches have been acquiring the ability to disentangle style attributes from the contents of the input images, producing better style transfer results~\cite{taigman2016unsupervised,pondenkandath2019historical}. Geometry-aware synthesizing methods~\cite{zhan2019spatial,zhan2019ga} have been successfully applied on scene text images, but cursive words are not considered.

Concerning the generation of handwritten text, such approaches have been mainly used for synthesising Chinese ideograms~\cite{lyu2017auto,tian2017zi2zi,chang2018generating,jiang2018w,wu2020calligan} and glyphs~\cite{azadi2018multi}. However, they are restricted to a predefined set of content classes. The incapability to generate out of vocabulary (OOV) text limits its practical application. Few works can actually deal with OOV words. First, in the work by Alonso \emph{et al.}~\cite{alonso2019adversarial}, the generation of handwritten word samples is conditioned by character sequences, but it suffers from the mode collapse problem, hindering the diversity of the generated images. Second, Fogel~\emph{et al.}~\cite{fogel2020scrabblegan} generate handwritten word by assembling the images generated by its characters, but the generated characters have the same receptive field width, which can make the generated words look unrealistic. Third, Mayr~\emph{et al.}~\cite{mayr2020spatio} propose a conversion model to approximate online handwriting from offline data and then apply style transfer method to online data, so that offline handwritten text images could be generated by leveraging online handwriting synthesizer. However, this method highly depends on the performance of the conversion model and needs online data to train.  Techniques like FUNIT~\cite{liu2019few}, able to transfer unseen target styles to the content generated images could be beneficial for this limitation. In particular, the use of Adaptive Instance Normalization (AdaIN) layers, proposed in~\cite{huang2017arbitrary}, shall allow to align both textual content and style attributes within the generative process. 

Summarizing, state-of-the-art generative methods are still unable to produce plausible yet diverse images of whatever handwritten word automatically. In this paper we propose to condition a generative model for handwritten words with unconstrained text sequences and stylistic typographic attributes, so that we are able to generate any word with a great diversity over the produced results. 

%%%%%%%%%%%%%%%%%%%%%%%%%%%%%%%%%%%%%%%%%%%%%%%%%%%%%%%%%%%%%%%%%%%%%%%%%
%%%%%%%%%%%%%%%%%%%%%%%%%%%%%%%%%%%%%%%%%%%%%%%%%%%%%%%%%%%%%%%%%%%%%%%%%
\section{Conditioned Handwritten Generation}
%%%%%%%%%%%%%%%%%%%%%%%%%%%%%%%%%%%%%%%%%%%%%%%%%%%%%%%%%%%%%%%%%%%%%%%%%
%%%%%%%%%%%%%%%%%%%%%%%%%%%%%%%%%%%%%%%%%%%%%%%%%%%%%%%%%%%%%%%%%%%%%%%%%
\subsection{Problem Formulation}
Let $\{\mathcal{X},\mathcal{Y},\mathcal{W}\}$ be a multi-writer handwritten word dataset, containing grayscale word images $\mathcal{X}$, their corresponding transcription strings $\mathcal{Y}$ and their writer identifiers $\mathcal{W}=\{w_i\}_{i=1}^N$. Let $X_i = \{x_{w_i,j}\}_{j=1}^K \subset \mathcal{X}$ be a subset of $K$ randomly sampled handwritten word images from the same given writer $w_i \in \mathcal{W}$. Let $\mathcal{A}$ be the alphabet containing the allowed characters (letters, digits, punctuation signs, etc.), $\mathcal{A}^l$ being all the possible text strings with length $l$. Given a set of images $X_i$ as a few-shot example of the calligraphic style attributes for writer $w_i$ on the one hand, and given a textual content provided by any text string $t \in \mathcal{A}^l$ on the other hand; the proposed generative model has the ability to combine both sources of information. It has the objective to yield a handwritten word image having textual content equal to $t$ and sharing calligraphic style attributes with writer $w_i$. Following this formulation, the generative model $H$ is defined as
\begin{equation}
    \bar{x} = H\left(t, X_i\right) = H\left(t, \left\{x_1, \ldots, x_K\right\}\right),
\end{equation}
where $\bar{x}$ is the artificially generated handwritten word image with the desired properties. Moreover, we denote $\bar{\mathcal{X}}$ as the output distribution of the generative network $H$.

\begin{figure}[t!]
    \centering
    \includegraphics[width=\linewidth]{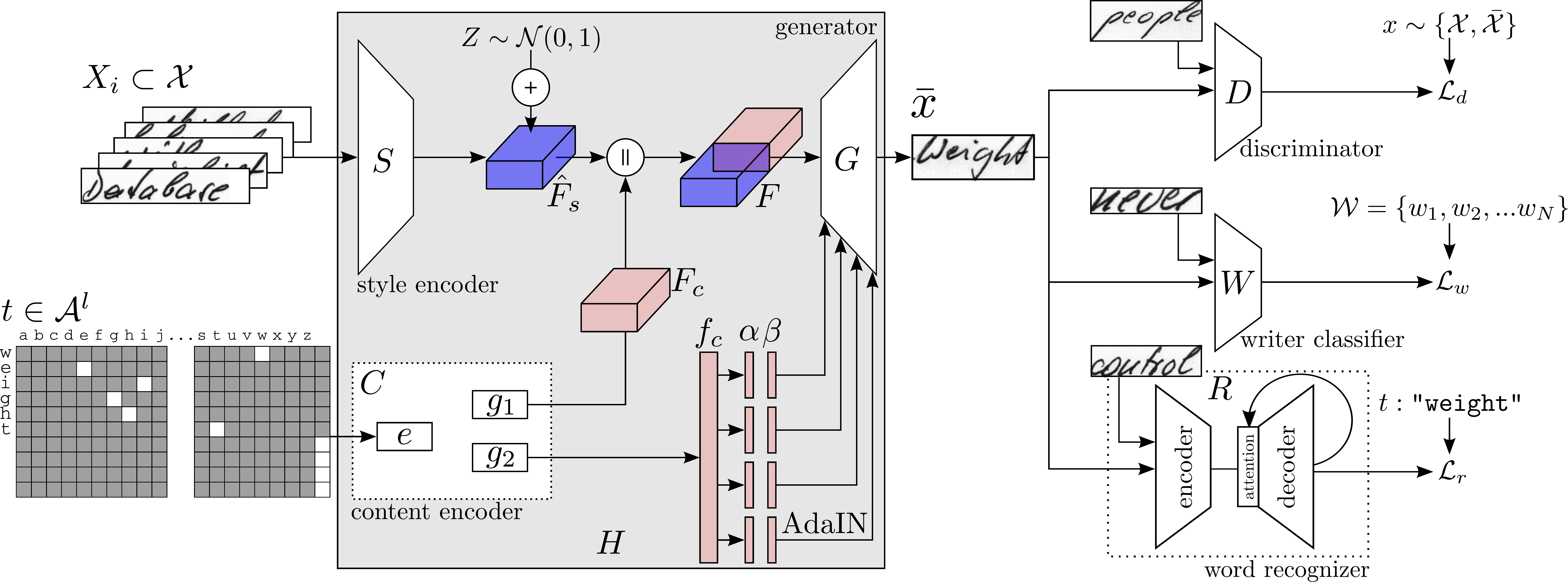}
    \caption{Architecture of the proposed handwriting generation model.}
    \label{fig:arch}
\end{figure}

The proposed architecture is divided in two main components. The generative network produces human-readable images conditioned to the combination of calligraphic style and textual content information. The second component are the learning objectives which guide the generative process towards producing images that look realistic; exhibiting a particular calligraphic style attributes; and having a specific textual content. Fig.~\ref{fig:arch} gives an overview of our model.
%%%%%%%%%%%%%%%%%%%%%%%%%%%%%%%%%%%%%%%%%%%%%%%%%%%%%%%%%%%%%%%%%%%%%%%%%
%%%%%%%%%%%%%%%%%%%%%%%%%%%%%%%%%%%%%%%%%%%%%%%%%%%%%%%%%%%%%%%%%%%%%%%%%
\subsection{Generative Network}
The proposed generative architecture $H$ consists of a calligraphic style encoder $S$, a textual content encoder $C$ and a conditioned image generator $G$. The overall calligraphic style of input images $X_i$ is disentangled from their individual textual contents, whereas the string $t$ provides the desired content.

\myparagraph{Calligraphic style encoding.} Given the set $X_i \subset \mathcal{X}$ of $K=15$ word images from the same writer $w_i$, the style encoder aims at extracting the calligraphic style attributes, \emph{i.e.}~slant, glyph shapes, stroke width, character roundness, ligatures etc. from the provided input samples. Specifically, our proposed network $S$
learns a style latent space mapping, in which the obtained style representations $F_s = S(X_i)$ are disentangled from the actual textual contents of the images $X_i$. The VGG-19-BN~\cite{simonyan2014very} architecture is used as the backbone of $S$. In order to process the input image set $X_i$, all the images are resized to have the same height $h$, padded to meet a maximum width $w$ and concatenated channel-wise to end up with a single tensor $h\times w\times K$. If we ask a human to write the same word several times, slight involuntary variations appear. In order to imitate this phenomenon, randomly choosing permutations of the subset $X_i$ will already produce such characteristic fluctuations. In addition, an additive noise $Z \sim \mathcal{N}(0, 1)$ is applied to the output latent space to obtain a subtly distorted feature representation $\hat{F_s} = F_s + Z$.

\myparagraph{Textual content encoding.} The textual content network $C$ is devoted to produce an encoding of the given text string $t$ that we want to artificially write. The proposed architecture outputs content features at two different levels. Low-level features encode the different characters that form a word and their spatial position within the string. A subsequent broader representation aims at guiding the whole word consistency. Formally, let $t\in \mathcal{A}^l$ be the input text string, character sequences shorter than $l$ are padded with the empty symbol $\varepsilon$. Let us define a character-wise embedding function $e \colon \mathcal{A} \to \mathbb{R}^n$. The first step of the content encoding stage embeds with a linear layer each character $c\in t$, represented by a one-hot vector, into a character-wise latent space. Then, the architecture is divided into two branches.

\emph{Character-wise encoding:} Let $g_1 \colon \mathbb{R}^n \to \mathbb{R}^m$ be a Multi-Layer Perceptron (MLP). Each embedded character $e(c)$ is processed individually by $g_1$ and their results are later stacked together. 
In order to combine such representation with style features, we have to ensure that the content feature map meets the shape of $\hat{F_s}$. Each character embedding is repeated multiple times horizontally to coarsely align the content features with the visual ones extracted from the style network, and the tensor is finally vertically expanded. The two feature representations are concatenated to be fed to the generator $F = [\hat{F_s} \parallel F_c]$. Such a character-wise encoding enables the network to produce OOV words, \emph{i.e.} words that have never been seen during training.

\emph{Global string encoding:} Let $g_2 \colon \mathbb{R}^{l\cdot n} \to \mathbb{R}^{2p\cdot q}$ be another MLP aimed at obtaining a much broader and global string representation. The character embeddings $e(c)$ are concatenated into a large one-dimensional vector of size $l \cdot n$ that is then processed by $g_2$. Such global representation vector $f_c$ will be then injected into the generator splitted into $p$ pairs of parameters.

Both functions $g_1(\cdot)$ and $g_2(\cdot)$ make use of three fully-connected layers with ReLU activation functions and batch normalization~\cite{ioffe2016batchnorm}.

\myparagraph{Generator.} Let $F$ be the combination of the calligraphic style attributes and the textual content information character-wise; and $f_c$ the global textual encoding. The generator $G$ is composed of two residual blocks~\cite{huang2018multimodal} using the AdaIN as the normalization layer. Then, four convolutional modules with nearest neighbor up-sampling and a final $\tanh$ activation layer generates the output image $\bar{x}$. AdaIN is formally defined as
\begin{equation}
	    \operatorname{AdaIN}\left(z, \alpha, \beta\right) = \alpha \left(\frac{z-\mu\left(z\right)}{\sigma\left(z\right)}\right) + \beta,
\end{equation}
where $z \in F$, $\mu$ and $\sigma$ are the channel-wise mean and standard deviations. The global content information is injected four times ($p=4$) during the generative process by the AdaIN layers. Their parameters $\alpha$ and $\beta$ are obtained by splitting $f_c$ in four pairs. Hence, the generative network is defined as 
\begin{equation}
    \bar{x} = H\left(t,X_i\right) = G\left( C\left(t\right), S\left(X_i\right)\right) = G\left( g_1\left(\hat{t}\right), g_2\left(\hat{t}\right), S\left(X_i\right)\right),
\end{equation}
where $\hat{t} = \left[e(c) ; \forall c\in t\right]$ is the encoding of the string $t$ character by character.

%%%%%%%%%%%%%%%%%%%%%%%%%%%%%%%%%%%%%%%%%%%%%%%%%%%%%%%%%%%%%%%%%%%%%%%%%
%%%%%%%%%%%%%%%%%%%%%%%%%%%%%%%%%%%%%%%%%%%%%%%%%%%%%%%%%%%%%%%%%%%%%%%%%
\subsection{Learning Objectives}
We propose to combine three complementary learning objectives: a discriminative loss, a style classification loss and a textual content loss. Each one of these losses aim at enforcing different properties of the desired generated image $\bar{x}$.

\myparagraph{Discriminative Loss.} Following the paradigm of GANs~\cite{goodfellow2014generative}, we make use of a discriminative model $D$ to estimate the probability that samples come from a real source, \emph{i.e.} training data $\mathcal{X}$, or belong to the artificially generated distribution $\bar{\mathcal{X}}$. Taking the generative network $H$ and the discriminator $D$, this setting corresponds to a $\min \max$ optimization problem. The proposed discriminator $D$ starts with a convolutional layer, followed by six residual blocks with LeakyReLU activations and average poolings. A final binary classification layer is used to discern between fake and real images. Thus, the discriminative loss only controls that the general visual appearance of the generated image looks realistic. However, it does not take into consideration neither the calligraphic styles nor the textual contents. This loss is formally defined as
\begin{equation}
    \mathcal{L}_d\left(H,D\right) = \mathbb{E}_{x \sim \mathcal{X}} \left[ \log\left(D\left(x\right)\right) \right] + \mathbb{E}_{\bar{x} \sim \bar{\mathcal{X}}}\left[\log\left(1-D\left(\bar{x}\right)\right)\right].
    \label{e:discriminator}
\end{equation}

\myparagraph{Style Loss.} When generating realistic handwritten word images, encoding information related to calligraphic styles not only provides diversity on the generated samples, but also prevents the mode collapse problem. Calligraphy is a strong identifier of different writers. In that sense, the proposed style loss guides the generative network $H$ to generate samples conditioned to a particular writing style by means of a writer classifier $W$. Given a handwritten word image, $W$ tries to identify the writer $w_i \in \mathcal{W}$ who produced it. The writer classifier $W$ follows the same architecture of the discriminator $D$ with a final classification MLP with the amount of writers in our training dataset. The classifier $W$ is only optimized with real samples drawn from $\mathcal{X}$, but it is used to guide the generation of the synthetic ones. We use the cross entropy loss, formally defined as
\begin{equation}
    \mathcal{L}_w\left(H, W\right) = - \mathbb{E}_{x \sim \left\{ \mathcal{X},\bar{\mathcal{X}} \right\}} \left[ \sum_{i=1}^{\left|\mathcal{W}\right|} w_i \log\left(\hat{w}_i\right) \right],
    \label{e:style}
\end{equation}
where $\hat{w} = W(x)$ is the predicted probability distribution over writers in $\mathcal{W}$ and $w_i$ the real writer distribution. Generated samples should be classified as the writer $w_i$ used to construct the input style conditioning image set $X_i$.

\begin{figure}[t!]
    \centering
    \includegraphics[width=.7\linewidth]{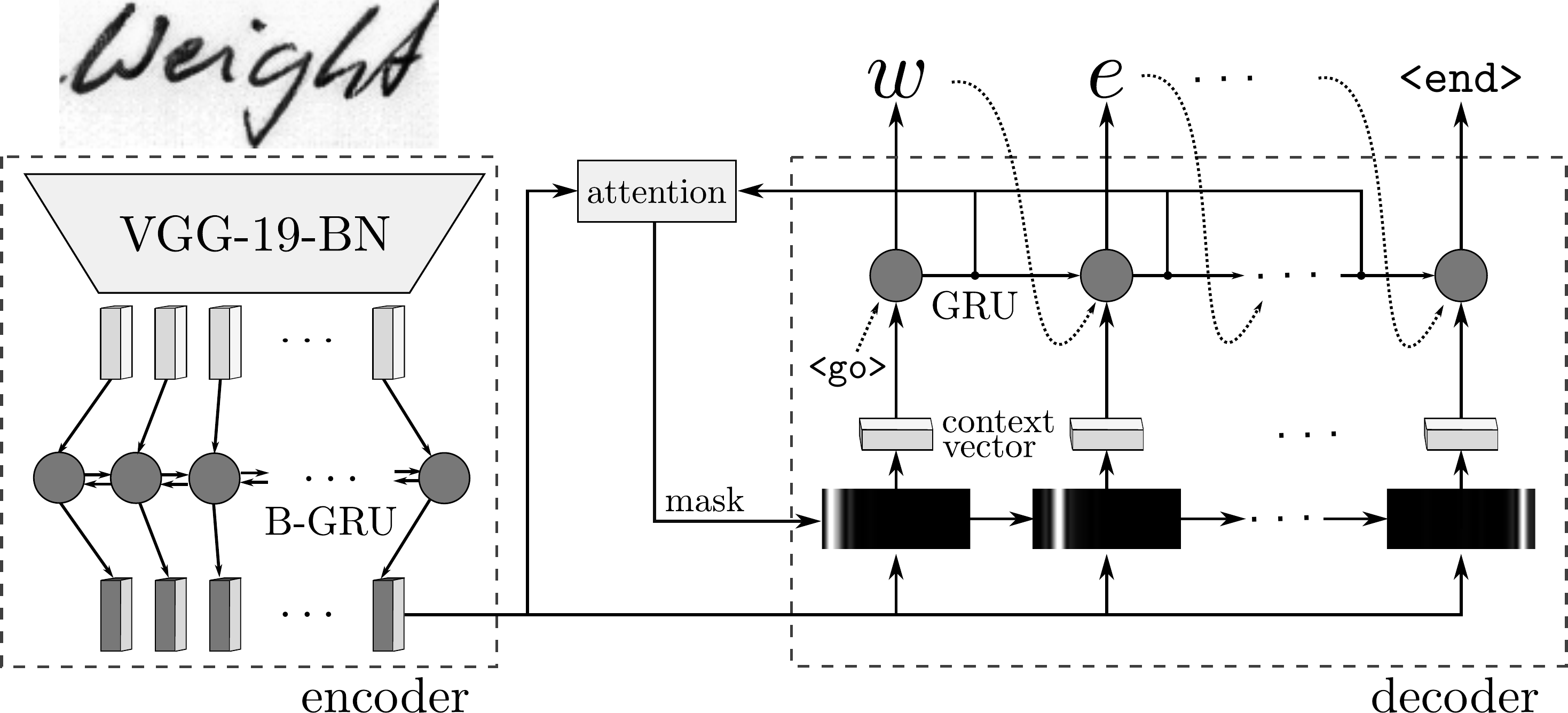}
    \caption{Architecture of the attention-based sequence-to-sequence handwritten word recognizer $R$.}
    \label{fig:arch2}
\end{figure}

\myparagraph{Content Loss.} A final handwritten word recognizer network $R$ is used to guide our generator towards producing synthetic word images with a specific textual content. We implemented a state-of-the-art sequence-to-sequence model~\cite{michael2019evaluating} for handwritten word recognition to examine whether the produced images $\bar{x}$ are actually decoded as the string $t$. The recognizer, depicted in Fig.~\ref{fig:arch2}, consists of an encoder and a decoder coupled with an attention mechanism. Handwritten word images are processed by the encoder and high-level feature representations are obtained. A VGG-19-BN~\cite{simonyan2014very} architecture followed by a two-layered Bi-directional Gated Recurrent Unit (B-GRU)~\cite{chung2014empirical} is used as the encoder network. The decoder is a one-directional RNN that outputs character by character predictions at each time step. The attention mechanism dynamically aligns context features from each time step of the decoder with high-level features from the encoder, hopefully corresponding to the next character to decode. The Kullback-Leibler divergence loss is used as the recognition loss at each time step. This is formally defined as
\begin{equation}
    \mathcal{L}_r\left(H, R\right) = - \mathbb{E}_{x \sim \left\{ \mathcal{X},\bar{\mathcal{X}} \right\}} \left[ \sum_{i=0}^{l} \sum_{j=0}^{\left|\mathcal{A}\right|}  t_{i,j} \log\left(\frac{t_{i,j}}{\hat{t}_{i,j}}\right) \right],
    \label{e:content}
\end{equation}
where $\hat{t} = R(x)$; $\hat{t}_{i}$ being the $i$-th decoded character probability distribution by the word recognizer, $\hat{t}_{i,j}$being the probability of $j$-th symbol in $\mathcal{A}$ for $\hat{t}_{i}$, and $t_{i,j}$ being the real probability corresponding to $\hat{t}_{i,j}$. The empty symbol $\varepsilon$ is ignored in the loss computation; $t_i$ denotes the $i$-th character on the input text $t$.

%%%%%%%%%%%%%%%%%%%%%%%%%%%%%%%%%%%%%%%%%%%%%%%%%%%%%%%%%%%%%%%%%%%%%%%%%
%%%%%%%%%%%%%%%%%%%%%%%%%%%%%%%%%%%%%%%%%%%%%%%%%%%%%%%%%%%%%%%%%%%%%%%%%
\subsection{End-to-end Training}
Overall, the whole architecture is trained end to end with the combination of the three proposed loss functions
\begin{equation}
    \mathcal{L}(H, D, W, R) = \mathcal{L}_d(H, D) + \mathcal{L}_w(H, W) + \mathcal{L}_r(H, R),
    \label{e:final}
\end{equation}
\begin{equation}
    \min_{H,W,R} \max_D \mathcal{L}(H, D, W, R).
\end{equation}
Algorithm~\ref{alg:train} presents the training strategy that has been followed in this work. $\Gamma(\cdot)$ denotes the optimizer function. Note that the parameter optimization is performed in two steps. First, the discriminative loss is computed using both real and generated samples (line~\ref{alg:line:D}). The style and content losses are computed by just providing real data (line~\ref{alg:line:wr}). 
Even though $W$ and $D$ are optimized using only real data and, therefore, they could be pre-trained independently from the generative network $H$, we obtained better results by initializing all the networks from scratch and jointly training them altogether. The network parameters $\Theta_D$ are optimized by gradient ascent following the GAN paradigm whereas the parameters $\Theta_W$ and $\Theta_R$ are optimized by gradient descent. Finally, the overall generator loss is computed following Equation~\ref{e:final} where only the generator parameters $\Theta_H$ are optimized (line~\ref{alg:line:H}).

\begin{algorithm}[t!]
\hspace*{\algorithmicindent} \textbf{Input:} Input data $\{\mathcal{X}, \mathcal{Y}, \mathcal{W}\}$; alphabet $\mathcal{A}$; max training iterations $T$ \\ 
\hspace*{\algorithmicindent} \textbf{Output: } Networks parameters $\{\Theta_{H}, \Theta_{D}, \Theta_{W}, \Theta_{R}\}$.
\begin{algorithmic}[1]
\Repeat
\State Get style and content mini-batches $\{X_i, w_i\}_{i=1}^{N_B}$ and $\{t^i\}_{i=1}^{N_B}$
\State $\mathcal{L}_d \leftarrow $ Eq.~\ref{e:discriminator} \label{alg:line:D} \algorithmiccomment{Real and generated samples $x \sim \{\mathcal{X}, \bar{\mathcal{X}}\}$}
\State $\mathcal{L}_{w,r} \leftarrow $ Eq.~\ref{e:style} + Eq.~\ref{e:content} \label{alg:line:wr} \algorithmiccomment{Real samples $x \sim \mathcal{X}$}
\State $\Theta_D \leftarrow \Theta_D + \Gamma(\nabla_{\Theta_D}\mathcal{L}_d)$
\State $\Theta_{W,R} \leftarrow \Theta_{W,R} - \Gamma(\nabla_{\Theta_{W,R}}\mathcal{L}_{w,d})$
\State $\mathcal{L} \leftarrow $ Eq.~\ref{e:final} \algorithmiccomment{Generated samples $x \sim \bar{\mathcal{X}}$}
\State $\Theta_H \leftarrow \Theta_H - \Gamma(\nabla_{\Theta_{H}}\mathcal{L})$ \label{alg:line:H}
\Until{Max training iterations $T$}
\end{algorithmic}
\caption{Training algorithm for the proposed model.} \label{alg:train}
\end{algorithm}

%%%%%%%%%%%%%%%%%%%%%%%%%%%%%%%%%%%%%%%%%%%%%%%%%%%%%%%%%%%%%%%%%%%%%%%%%
%%%%%%%%%%%%%%%%%%%%%%%%%%%%%%%%%%%%%%%%%%%%%%%%%%%%%%%%%%%%%%%%%%%%%%%%%
\section{Experiments}
To carry out the different experiments, we have used a subset of the IAM corpus~\cite{marti2002iam} as our multi-writer handwritten dataset $\{\mathcal{X},\mathcal{Y},\mathcal{W}\}$. It consists of $62,857$ handwritten word snippets, written by 500 different individuals. Each word image has its associated writer and transcription metadata. A test subset of 160 writers has been kept apart during training to check whether the generative model is able to cope with unseen calligraphic styles. We have also used a subset of $22,500$ unique English words from the Brown~\cite{bird2009natural} corpus as the source of strings for the content input. A test set of 400 unique words, disjoint from the IAM transcriptions has been used to test the performance when producing OOV words. To quantitatively measure the image quality, diversity and the ability to transfer style attributes of the proposed approach we will use the Fr{\'e}chet Inception Distance (FID)~\cite{heusel2017gans,borji2019pros}, measuring the distance between the Inception-v3 activation distributions for generated $\bar{\mathcal{X}}$ and real samples $\mathcal{X}$ for each writer $w_i$ separately, and finally averaging them. Inception features, trained over ImageNet data, have not been designed to discern between different handwriting images. Although this measure might not be ideal to evaluate our specific case, it will still serve as an indication of the similarity between generated and real images.

%%%%%%%%%%%%%%%%%%%%%%%%%%%%%%%%%%%%%%%%%%%%%%%%%%%%%%%%%%%%%%%%%%%%%%%%%
%%%%%%%%%%%%%%%%%%%%%%%%%%%%%%%%%%%%%%%%%%%%%%%%%%%%%%%%%%%%%%%%%%%%%%%%%
\begingroup
\setlength{\tabcolsep}{1pt} % Default value: 6pt
\renewcommand{\arraystretch}{1.25} % Default value: 1
\begin{figure}[t!]
    \centering
    \begin{tabular}{l@{\hskip 6pt}ccccccc}
    a) \tiny{IV-S}&\includegraphics[width=0.12\linewidth, valign=c]{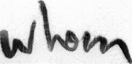}&
    \includegraphics[width=0.12\linewidth, valign=c]{images/in_tr/013-39_twice-trice.png}&
    \includegraphics[width=0.12\linewidth, valign=c]{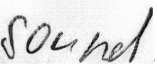}&
    \includegraphics[width=0.12\linewidth, valign=c]{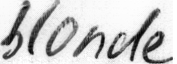}&
    \includegraphics[width=0.12\linewidth, valign=c]{images/in_tr/024-68_mouth-month.png}&
    \includegraphics[width=0.12\linewidth, valign=c]{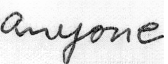}&
    \includegraphics[width=0.12\linewidth, valign=c]{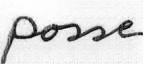}\\
    b) \tiny{IV-U}&\includegraphics[width=0.12\linewidth, valign=c]{images/in_te/168-19_sound-sound.png}&
    \includegraphics[width=0.12\linewidth, valign=c]{images/in_te/180-21_anyone-anyone.png}&
    \includegraphics[width=0.12\linewidth, valign=c]{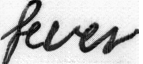}&
    \includegraphics[width=0.12\linewidth, valign=c]{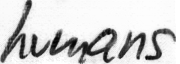}&
    \includegraphics[width=0.12\linewidth, valign=c]{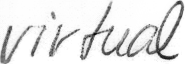}&
    \includegraphics[width=0.12\linewidth, valign=c]{images/in_te/198-40_Having-Having.png}&
    \includegraphics[width=0.12\linewidth, valign=c]{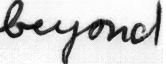}\\
    c) \tiny{OOV-S}&\includegraphics[width=0.12\linewidth, valign=c]{images/oo_tr/000-30_reviews-reviems.png}&
    \includegraphics[width=0.12\linewidth, valign=c]{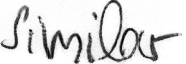}&
    \includegraphics[width=0.12\linewidth, valign=c]{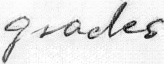}&
    \includegraphics[width=0.12\linewidth, valign=c]{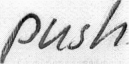}&
    \includegraphics[width=0.12\linewidth, valign=c]{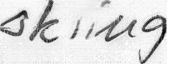}&
    \includegraphics[width=0.12\linewidth, valign=c]{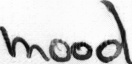}&
    \includegraphics[width=0.12\linewidth, valign=c]{images/oo_tr/134-78_monkey-monkey.png}\\
    d) \tiny{OOV-U}&\includegraphics[width=0.12\linewidth, valign=c]{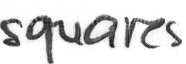}&
    \includegraphics[width=0.12\linewidth, valign=c]{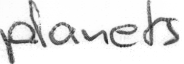}&
    \includegraphics[width=0.12\linewidth, valign=c]{images/oo_te/173-72_push-push.png}&
    \includegraphics[width=0.12\linewidth, valign=c]{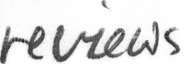}&
    \includegraphics[width=0.12\linewidth, valign=c]{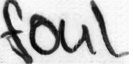}&
    \includegraphics[width=0.12\linewidth, valign=c]{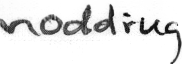}&
    \includegraphics[width=0.12\linewidth, valign=c]{images/oo_te/545-65_Flower-Flower.png}\\
    \end{tabular}
    \caption{Word image generation. a) In-Vocabulary (IV) words and seen (S) styles; b) In-Vocabulary (IV) words and unseen (U) styles; c) Out-of-Vocabulary (OOV) words and seen (S) styles and d) Out-of-Vocabulary (OOV) words and unseen (U) styles.}
    \label{fig:sample}
\end{figure}
\endgroup

\subsection{Generating Handwritten Word Images}
We present in Fig.~\ref{fig:sample} an illustrative selection of generated handwritten words. We appreciate the realistic and diverse aspect of the produced images. Qualitatively, we observe that the proposed approach is able to yield satisfactory results even when dealing with both words and calligraphic styles never seen during training. But, when analyzing the different experimental settings in Table~\ref{tab:4_scenario}, we appreciate that the FID measure slightly degrades when either we input an OOV word or a style never seen during training. Nevertheless, the reached FID measures in all four settings satisfactorily compare with the baseline achieved by real data.

\begin{table}[t!]
    \caption{FID between generated images and real images of corresponding set.}
    \label{tab:4_scenario}
    \centering
    \begin{tabular}{lccccc}
    \toprule
    &Real images & IV-S & IV-U & OOV-S & OOV-U\\
    \midrule
    FID & $90.43$ & $120.07$ & $124.30$& $125.87$ &$130.68$\\
    \bottomrule
    \end{tabular}
\end{table}
\begin{figure}[h!t]
    \centering
    \includegraphics[width=\linewidth]{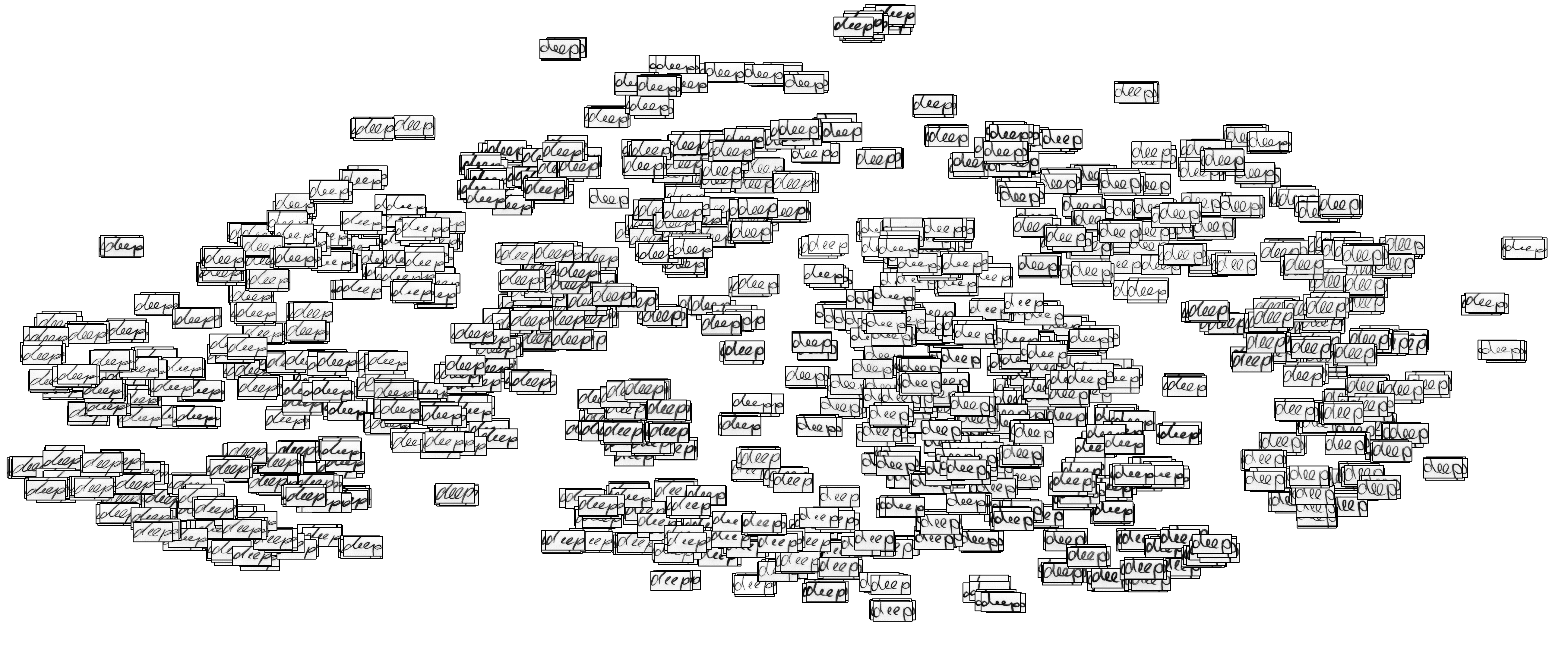}
    \caption{t-SNE embedding visualization of $2.500$ generated instances of the word \texttt{"deep"}.}
    \label{fig:tsne}
\end{figure}

In order to show the ability of the proposed method to produce a diverse set of generated images, we present in Fig.~\ref{fig:tsne} a t-SNE~\cite{maaten2008visualizing} visualization of different instances produced with a fixed textual content while varying the calligraphic style inputs. Different clusters corresponding to particular slants, stroke widths, character roundnesses, ligatures and cursive writings are observed.

\begin{figure}[t!]
    \centering
    \begin{tabular}{c@{\hskip 8pt} ccccccc}
    \toprule
        && \multicolumn{6}{c}{\textbf{Style Images}}\\
        \cmidrule{3-8}
        && 
        \includegraphics[width=.12\linewidth, valign=c]{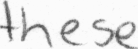} &
        \includegraphics[width=.12\linewidth, valign=c]{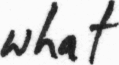} &
        \includegraphics[width=.12\linewidth, valign=c]{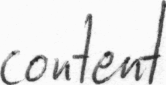} &
        \includegraphics[width=.12\linewidth,, valign=c]{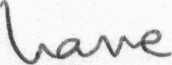} &
        \includegraphics[width=.12\linewidth, valign=c]{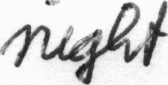} &
        \includegraphics[width=.12\linewidth, valign=c]{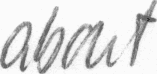} \\
        \midrule
    \multirow{12}{*}{\raisebox{-8pt}{\rotatebox[origin=r]{90}{\textbf{Textual Content}}}} & \textbf{FUNIT} &
        \multirow{2}{*}{\includegraphics[width=.12\linewidth, valign=c]{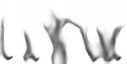}} &
        \multirow{2}{*}{\includegraphics[width=.12\linewidth, valign=c]{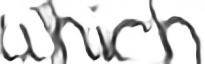}} &
        \multirow{2}{*}{\includegraphics[width=.12\linewidth, valign=c]{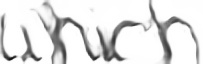}} &
        \multirow{2}{*}{\includegraphics[width=.12\linewidth, valign=c]{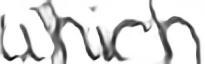}} &
        \multirow{2}{*}{\includegraphics[width=.12\linewidth, valign=c]{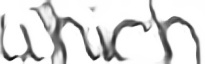}} &
        \multirow{2}{*}{\includegraphics[width=.12\linewidth, valign=c]{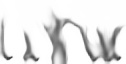}} \\
       &  \includegraphics[height=0.3cm, valign=c]{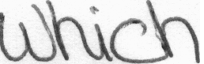} &&&&&\\
        \cmidrule{3-8}
        & \textbf{ours} &
        \multirow{2}{*}{\includegraphics[width=.12\linewidth, valign=c]{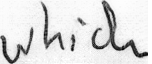}} &
        \multirow{2}{*}{\includegraphics[width=.12\linewidth, valign=c]{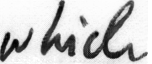}} &
        \multirow{2}{*}{\includegraphics[width=.12\linewidth, valign=c]{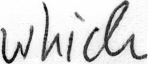}} &
        \multirow{2}{*}{\includegraphics[width=.12\linewidth, valign=c]{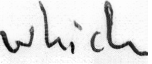}} &
        \multirow{2}{*}{\includegraphics[width=.12\linewidth, valign=c]{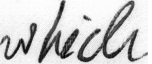}} &
        \multirow{2}{*}{\includegraphics[width=.12\linewidth, valign=c]{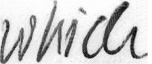}} \\
        & \texttt{"which"} &&&&&\\
        \cmidrule{2-8}
        & \textbf{FUNIT} &
        \multirow{2}{*}{\includegraphics[width=.12\linewidth, valign=c]{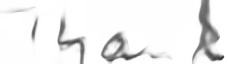}} &
        \multirow{2}{*}{\includegraphics[width=.12\linewidth, valign=c]{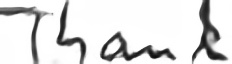}} &
        \multirow{2}{*}{\includegraphics[width=.12\linewidth, valign=c]{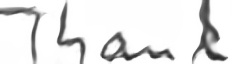}} &
        \multirow{2}{*}{\includegraphics[width=.12\linewidth, valign=c]{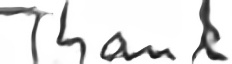}} &
        \multirow{2}{*}{\includegraphics[width=.12\linewidth, valign=c]{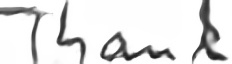}} &
        \multirow{2}{*}{\includegraphics[width=.12\linewidth, valign=c]{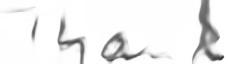}} \\
        & \includegraphics[height=0.3cm, valign=c]{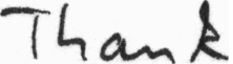} &&&&&\\
        \cmidrule{3-8}
        & \textbf{ours} &
        \multirow{2}{*}{\includegraphics[width=.12\linewidth, valign=c]{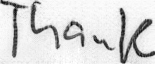}} &
        \multirow{2}{*}{\includegraphics[width=.12\linewidth, valign=c]{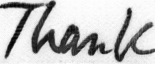}} &
        \multirow{2}{*}{\includegraphics[width=.12\linewidth, valign=c]{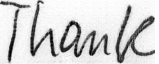}} &
        \multirow{2}{*}{\includegraphics[width=.12\linewidth, valign=c]{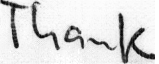}} &
        \multirow{2}{*}{\includegraphics[width=.12\linewidth, valign=c]{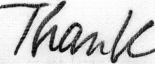}} &
        \multirow{2}{*}{\includegraphics[width=.12\linewidth, valign=c]{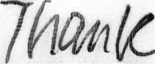}} \\
        & \texttt{"Thank"} &&&&&\\
        \cmidrule{2-8}
        & \textbf{FUNIT} &
        \multirow{2}{*}{\includegraphics[width=.12\linewidth, valign=c]{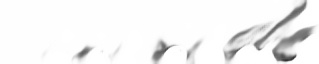}} &
        \multirow{2}{*}{\includegraphics[width=.12\linewidth, valign=c]{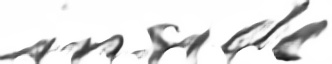}} &
        \multirow{2}{*}{\includegraphics[width=.12\linewidth, valign=c]{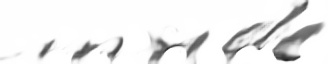}} &
        \multirow{2}{*}{\includegraphics[width=.12\linewidth, valign=c]{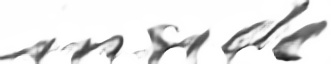}} &
        \multirow{2}{*}{\includegraphics[width=.12\linewidth, valign=c]{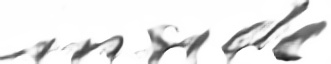}} &
        \multirow{2}{*}{\includegraphics[width=.12\linewidth, valign=c]{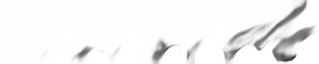}} \\
        & \includegraphics[height=0.3cm, valign=c]{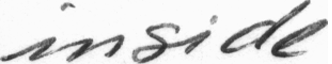} &&&&&\\
        \cmidrule{3-8}
        & \textbf{ours} &
        \multirow{2}{*}{\includegraphics[width=.12\linewidth, valign=c]{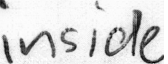}} &
        \multirow{2}{*}{\includegraphics[width=.12\linewidth, valign=c]{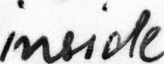}} &
        \multirow{2}{*}{\includegraphics[width=.12\linewidth, valign=c]{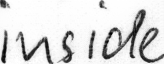}} &
        \multirow{2}{*}{\includegraphics[width=.12\linewidth, valign=c]{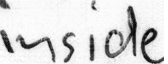}} &
        \multirow{2}{*}{\includegraphics[width=.12\linewidth, valign=c]{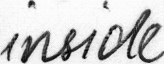}} &
        \multirow{2}{*}{\includegraphics[width=.12\linewidth, valign=c]{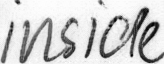}} \\
        & \texttt{"inside"} &&&&&\\
        \bottomrule
    \end{tabular}
    \caption{Comparison of handwritten word generation with FUNIT~\cite{liu2019few}.}
    \label{tab:funit}
\end{figure}

To further demonstrate the ability of the proposed approach to coalesce content and style information into the generated handwritten word images, we compare in Fig.~\ref{tab:funit} our produced results with the outcomes of the state-of-the-art approach FUNIT~\cite{liu2019few}. Being an image-to-image translation method, FUNIT starts with a content image and then injects the style attributes derived from a second sample image. Although FUNIT performs well for natural scene images, it is clear that such kind of approaches do not apply well for the specific case of handwritten words. Starting with a content image instead of a text string confines the generative process to the shapes of the initial drawing. When infusing the style features, the FUNIT method is only able to deform the stroke textures, often resulting in extremely distorted words. Conversely, our proposed generative process is able to produce realistic and diverse word samples given a content text string and a calligraphic style example. We observe how for the different produced versions of the same word, the proposed approach is able to
change style attributes as stroke width or slant, to produce both cursive words, where all characters are connected through ligatures as well as disconnected writings, and even render the same characters differently, \emph{e.g.} note the characters \texttt{n} or \texttt{s} in \texttt{"Thank"} or \texttt{"inside"} respectively.

%%%%%%%%%%%%%%%%%%%%%%%%%%%%%%%%%%%%%%%%%%%%%%%%%%%%%%%%%%%%%%%%%%%%%%%%%
%%%%%%%%%%%%%%%%%%%%%%%%%%%%%%%%%%%%%%%%%%%%%%%%%%%%%%%%%%%%%%%%%%%%%%%%%
\subsection{Latent Space Interpolations}

\begin{figure}[t!]
    \centering
    \resizebox{\linewidth}{!}{
    \begin{tabular}{rccccccccccc}
    & $w_A$ & & & & & & & & & & $w_B$\\ \midrule
    \tiny{Real} & \includegraphics[width=0.07\linewidth,valign=c]{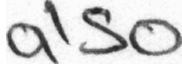} &
    & & & & & & & & & \includegraphics[width=0.07\linewidth,valign=c]{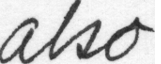} \\
    \tiny{Generated}&\includegraphics[width=0.07\linewidth,valign=c]{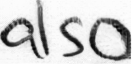} & 
    \includegraphics[width=0.07\linewidth,valign=c]{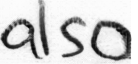} & 
    \includegraphics[width=0.07\linewidth,valign=c]{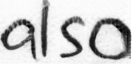} &
    \includegraphics[width=0.07\linewidth,valign=c]{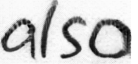} &
    \includegraphics[width=0.07\linewidth,valign=c]{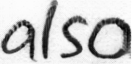} &
    \includegraphics[width=0.07\linewidth,valign=c]{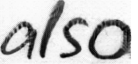} &
    \includegraphics[width=0.07\linewidth,valign=c]{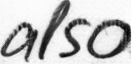} &
    \includegraphics[width=0.07\linewidth,valign=c]{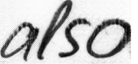} &
    \includegraphics[width=0.07\linewidth,valign=c]{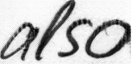} &
    \includegraphics[width=0.07\linewidth,valign=c]{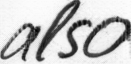} &
    \includegraphics[width=0.07\linewidth,valign=c]{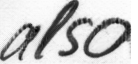} \\
    \midrule
    \tiny{Real} & \includegraphics[width=0.07\linewidth,valign=c]{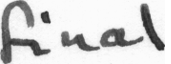} &
    & & & & & & & & & \includegraphics[width=0.07\linewidth,valign=c]{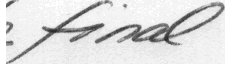} \\
    \tiny{Generated}&\includegraphics[width=0.07\linewidth,valign=c]{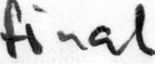} & 
    \includegraphics[width=0.07\linewidth,valign=c]{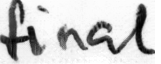} &
    \includegraphics[width=0.07\linewidth,valign=c]{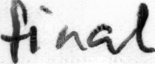} &
    \includegraphics[width=0.07\linewidth,valign=c]{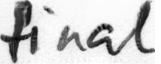} &
    \includegraphics[width=0.07\linewidth,valign=c]{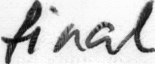} &
    \includegraphics[width=0.07\linewidth,valign=c]{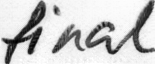} &
    \includegraphics[width=0.07\linewidth,valign=c]{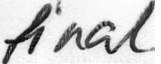} &
    \includegraphics[width=0.07\linewidth,valign=c]{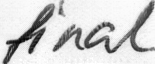} &
    \includegraphics[width=0.07\linewidth,valign=c]{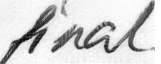} &
    \includegraphics[width=0.07\linewidth,valign=c]{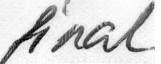} &
    \includegraphics[width=0.07\linewidth,valign=c]{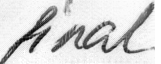} \\
    \midrule
    \tiny{Real} & \includegraphics[width=0.07\linewidth,valign=c]{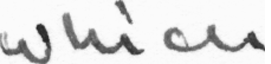} &
    & & & & & & & & & \includegraphics[width=0.07\linewidth,valign=c]{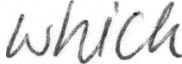} \\
    \tiny{Generated}&\includegraphics[width=0.07\linewidth,valign=c]{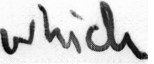} & 
    \includegraphics[width=0.07\linewidth,valign=c]{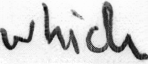} &
    \includegraphics[width=0.07\linewidth,valign=c]{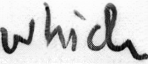} &
    \includegraphics[width=0.07\linewidth,valign=c]{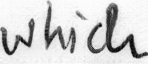} &
    \includegraphics[width=0.07\linewidth,valign=c]{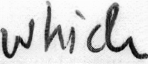} &
    \includegraphics[width=0.07\linewidth,valign=c]{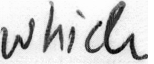} &
    \includegraphics[width=0.07\linewidth,valign=c]{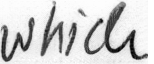} &
    \includegraphics[width=0.07\linewidth,valign=c]{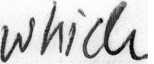} &
    \includegraphics[width=0.07\linewidth,valign=c]{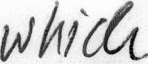} &
    \includegraphics[width=0.07\linewidth,valign=c]{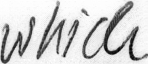} &
    \includegraphics[width=0.07\linewidth,valign=c]{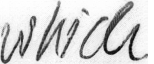} \\
    \midrule
    \tiny{Real} & \includegraphics[width=0.07\linewidth,valign=c]{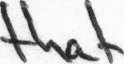} &
    & & & & & & & & & \includegraphics[width=0.07\linewidth,valign=c]{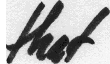} \\
    \tiny{Generated}&\includegraphics[width=0.07\linewidth,valign=c]{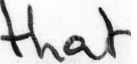} & 
    \includegraphics[width=0.07\linewidth,valign=c]{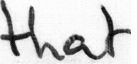} &
    \includegraphics[width=0.07\linewidth,valign=c]{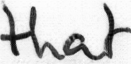} &
    \includegraphics[width=0.07\linewidth,valign=c]{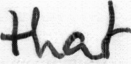} &
    \includegraphics[width=0.07\linewidth,valign=c]{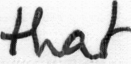} &
    \includegraphics[width=0.07\linewidth,valign=c]{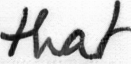} &
    \includegraphics[width=0.07\linewidth,valign=c]{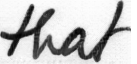} &
    \includegraphics[width=0.07\linewidth,valign=c]{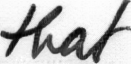} &
    \includegraphics[width=0.07\linewidth,valign=c]{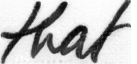} &
    \includegraphics[width=0.07\linewidth,valign=c]{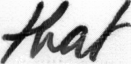} &
    \includegraphics[width=0.07\linewidth,valign=c]{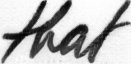} \\
    \midrule
    \tiny{Real} & \includegraphics[width=0.07\linewidth,valign=c]{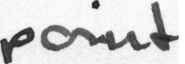} &
    & & & & & & & & & \includegraphics[width=0.07\linewidth,valign=c]{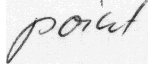} \\
    \tiny{Generated}&\includegraphics[width=0.07\linewidth,valign=c]{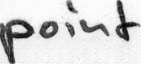} & 
    \includegraphics[width=0.07\linewidth,valign=c]{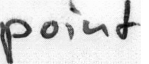} &
    \includegraphics[width=0.07\linewidth,valign=c]{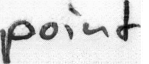} &
    \includegraphics[width=0.07\linewidth,valign=c]{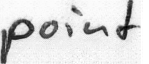} &
    \includegraphics[width=0.07\linewidth,valign=c]{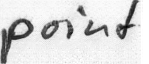} &
    \includegraphics[width=0.07\linewidth,valign=c]{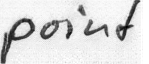} &
    \includegraphics[width=0.07\linewidth,valign=c]{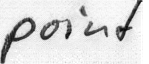} &
    \includegraphics[width=0.07\linewidth,valign=c]{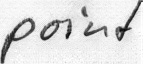} &
    \includegraphics[width=0.07\linewidth,valign=c]{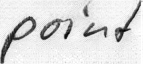} &
    \includegraphics[width=0.07\linewidth,valign=c]{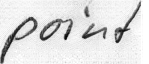} &
    \includegraphics[width=0.07\linewidth,valign=c]{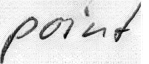} \\
    \midrule
    \tiny{Real} & \includegraphics[width=0.07\linewidth,valign=c]{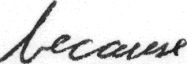} &
    & & & & & & & & & \includegraphics[width=0.07\linewidth,valign=c]{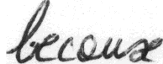} \\
    \tiny{Generated}&\includegraphics[width=0.07\linewidth,valign=c]{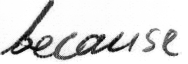} & 
    \includegraphics[width=0.07\linewidth,valign=c]{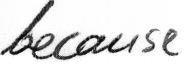} &
    \includegraphics[width=0.07\linewidth,valign=c]{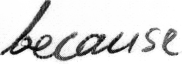} &
    \includegraphics[width=0.07\linewidth,valign=c]{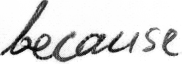} &
    \includegraphics[width=0.07\linewidth,valign=c]{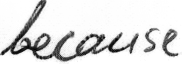} &
    \includegraphics[width=0.07\linewidth,valign=c]{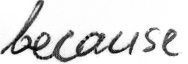} &
    \includegraphics[width=0.07\linewidth,valign=c]{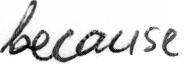} &
    \includegraphics[width=0.07\linewidth,valign=c]{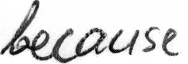} &
    \includegraphics[width=0.07\linewidth,valign=c]{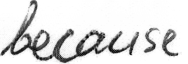} &
    \includegraphics[width=0.07\linewidth,valign=c]{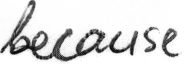} &
    \includegraphics[width=0.07\linewidth,valign=c]{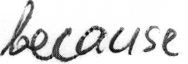} \\
    \bottomrule
\end{tabular}}
    \caption{Latent space interpolation between two calligraphic styles for different words while keeping contents fixed.}
    \label{fig:interpolation}
\end{figure}

The generator network $G$ learns to map feature points $F$ in the latent space to synthetic handwritten word images. Such latent space presents a structure worth exploring. We first interpolate in Fig.~\ref{fig:interpolation} between two different points $F_s^A$ and $F_s^B$ corresponding to two different calligraphic styles $w_A$ and $w_B$ while keeping the textual contents $t$ fixed. We observe how the generated images smoothly adjust from one style to another. Again note how individual characters evolve from one typography to another, \emph{e.g.} the \texttt{l} from \texttt{"also"}, or the \texttt{f} from \texttt{"final"}.

\begin{figure}[t!]
    \centering
    \begin{tabular}{cccccccccccc}
    \texttt{\scriptsize{"three"}}&
    \texttt{\scriptsize{"threw"}}&
    \texttt{\scriptsize{"shrew"}}&
    \texttt{\scriptsize{"shred"}}&
    \texttt{\scriptsize{"sired"}}&
    \texttt{\scriptsize{"fired"}}&
    \texttt{\scriptsize{"fined"}}&
    \texttt{\scriptsize{"firer"}}&
    \texttt{\scriptsize{"fiver"}}&
    \texttt{\scriptsize{"fever"}}&
    \texttt{\scriptsize{"sever"}}&
    \texttt{\scriptsize{"seven"}}\\
    \midrule
    \includegraphics[width=0.07\linewidth]{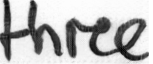}&
    \includegraphics[width=0.07\linewidth]{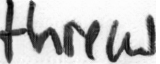}&
    \includegraphics[width=0.07\linewidth]{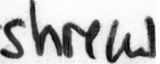}&
    \includegraphics[width=0.07\linewidth]{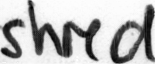}&
    \includegraphics[width=0.07\linewidth]{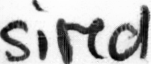}&
    \includegraphics[width=0.07\linewidth]{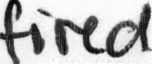}&
    \includegraphics[width=0.07\linewidth]{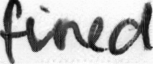}&
    \includegraphics[width=0.07\linewidth]{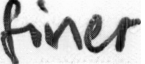}&
    \includegraphics[width=0.07\linewidth]{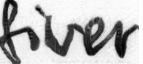}&
    \includegraphics[width=0.07\linewidth]{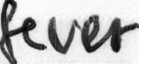}&
    \includegraphics[width=0.07\linewidth]{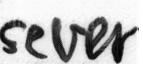}&
    \includegraphics[width=0.07\linewidth]{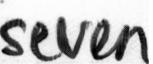}\\
    \includegraphics[width=0.07\linewidth]{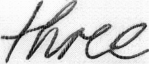}&
    \includegraphics[width=0.07\linewidth]{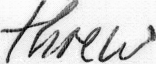}&
    \includegraphics[width=0.07\linewidth]{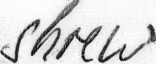}&
    \includegraphics[width=0.07\linewidth]{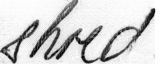}&
    \includegraphics[width=0.07\linewidth]{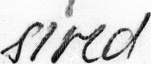}&
    \includegraphics[width=0.07\linewidth]{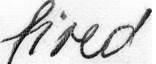}&
    \includegraphics[width=0.07\linewidth]{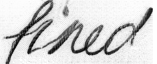}&
    \includegraphics[width=0.07\linewidth]{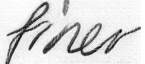}&
    \includegraphics[width=0.07\linewidth]{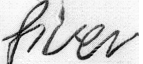}&
    \includegraphics[width=0.07\linewidth]{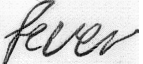}&
    \includegraphics[width=0.07\linewidth]{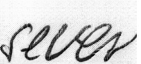}&
    \includegraphics[width=0.07\linewidth]{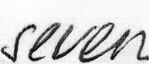}\\
    \includegraphics[width=0.07\linewidth]{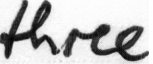}&
    \includegraphics[width=0.07\linewidth]{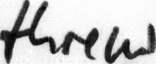}&
    \includegraphics[width=0.07\linewidth]{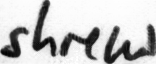}&
    \includegraphics[width=0.07\linewidth]{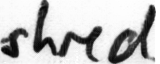}&
    \includegraphics[width=0.07\linewidth]{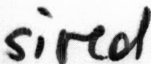}&
    \includegraphics[width=0.07\linewidth]{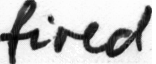}&
    \includegraphics[width=0.07\linewidth]{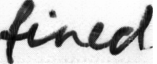}&
    \includegraphics[width=0.07\linewidth]{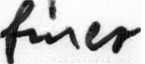}&
    \includegraphics[width=0.07\linewidth]{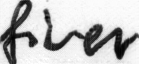}&
    \includegraphics[width=0.07\linewidth]{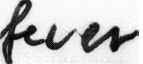}&
    \includegraphics[width=0.07\linewidth]{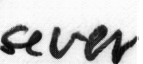}&
    \includegraphics[width=0.07\linewidth]{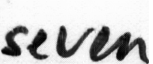}\\
    \includegraphics[width=0.07\linewidth]{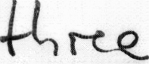}&
    \includegraphics[width=0.07\linewidth]{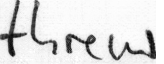}&
    \includegraphics[width=0.07\linewidth]{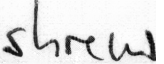}&
    \includegraphics[width=0.07\linewidth]{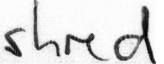}&
    \includegraphics[width=0.07\linewidth]{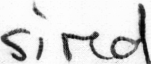}&
    \includegraphics[width=0.07\linewidth]{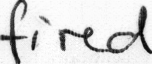}&
    \includegraphics[width=0.07\linewidth]{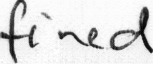}&
    \includegraphics[width=0.07\linewidth]{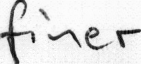}&
    \includegraphics[width=0.07\linewidth]{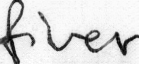}&
    \includegraphics[width=0.07\linewidth]{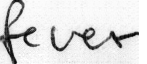}&
    \includegraphics[width=0.07\linewidth]{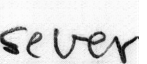}&
    \includegraphics[width=0.07\linewidth]{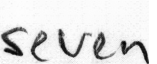}\\
    \includegraphics[width=0.07\linewidth]{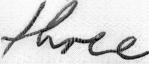}&
    \includegraphics[width=0.07\linewidth]{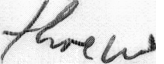}&
    \includegraphics[width=0.07\linewidth]{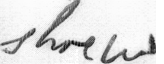}&
    \includegraphics[width=0.07\linewidth]{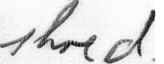}&
    \includegraphics[width=0.07\linewidth]{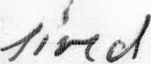}&
    \includegraphics[width=0.07\linewidth]{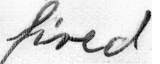}&
    \includegraphics[width=0.07\linewidth]{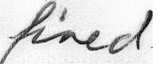}&
    \includegraphics[width=0.07\linewidth]{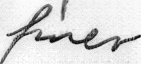}&
    \includegraphics[width=0.07\linewidth]{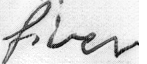}&
    \includegraphics[width=0.07\linewidth]{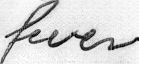}&
    \includegraphics[width=0.07\linewidth]{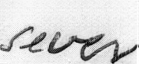}&
    \includegraphics[width=0.07\linewidth]{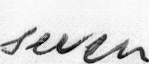}\\
    \bottomrule
    \end{tabular}
    \caption{Word ladder. From \texttt{"three"} to \texttt{"seven"} changing one character at a time, generated for five different calligraphic styles.}
    \label{fig:ladder}
\end{figure}

Contrary to the continuous nature of the style latent space, the original content space is discrete in nature. Instead of computing point-wise interpolations, we present in Fig.~\ref{fig:ladder} the obtained word images for different styles when following a ``word ladder'' puzzle game, \emph{i.e.} going from one word to another, one character difference at a time. Here we observe how different contents influence stylistic aspects. Usually \texttt{s} and \texttt{i} are disconnected when rendering the word \texttt{"sired"} but often appear with a ligature when jumping to the word \texttt{"fired"}.

%%%%%%%%%%%%%%%%%%%%%%%%%%%%%%%%%%%%%%%%%%%%%%%%%%%%%%%%%%%%%%%%%%%%%%%%%
%%%%%%%%%%%%%%%%%%%%%%%%%%%%%%%%%%%%%%%%%%%%%%%%%%%%%%%%%%%%%%%%%%%%%%%%%
\subsection{Impact of the Learning Objectives}

%%%%%%%%%%%%%%%%%%%%%%%%%%%%%%%%%%%%%%%%%%%%%%%%%%%%%%%%%%%%%%%%%%%%%%%%%
%%%%%%%%%%%%%%%%%%%%%%%%%%%%%%%%%%%%%%%%%%%%%%%%%%%%%%%%%%%%%%%%%%%%%%%%%
%  \begin{table}[!htb]
%     \centering
%     \caption{Content injection methods}
%     \label{tab:abla_content}
%     \begin{tabular}{cc}
%     \toprule
%         Method & FID\\
%         \midrule
%         AdaIN & $136.56$\\
%         Cat & $ 134.42$\\
%         \textbf{Both} & \textbf{$130.68$}\\
%     \bottomrule
%     \end{tabular}
% \end{table}
%%%%%%%%%%%%%%%%%%%%%%%%%%%%%%%%%%%%%%%%%%%%%%%%%%%%%%%%%%%%%%%%%%%%%%%%%
%%%%%%%%%%%%%%%%%%%%%%%%%%%%%%%%%%%%%%%%%%%%%%%%%%%%%%%%%%%%%%%%%%%%%%%%%
Along this paper, we have proposed to guide the generation process by three complementary goals. The discriminator loss $\mathcal{L}_d$ controlling the genuine appearance of the generated images $\bar{x}$. The writer classification loss $\mathcal{L}_w$ forcing $\bar{x}$ to mimic the calligraphic style of input images $X_i$. The recognition loss $\mathcal{L}_r$ guaranteeing that $\bar{x}$ is readable and conveys the exact text information $t$. We analyze in Table~\ref{tab:abla_loss} the effect of each learning objective. 

\begingroup
\setlength{\tabcolsep}{4pt} % Default value: 6pt
\begin{table}[t!]
    \centering
    \caption{Effect of each different learning objectives when generating the content $t=\texttt{"vision"}$ for different styles.}
    \label{tab:abla_loss}
    \begin{tabular}{c@{\hskip 4pt}c@{\hskip 4pt}c @{\hskip 8pt}ccccc}
    \toprule
        \multirow{2}{*}{$\mathcal{L}_d$} & \multirow{2}{*}{$\mathcal{L}_w$} & \multirow{2}{*}{$\mathcal{L}_r$} &
        \multirow{2}{*}{FID} & \multicolumn{4}{c}{\textbf{Style Images}} \\
        \cmidrule{5-8}
        &&&&\includegraphics[height=0.6cm,valign=c]{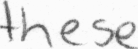} &
        \includegraphics[height=0.6cm,valign=c]{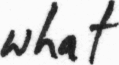} &
        \includegraphics[height=0.6cm,valign=c]{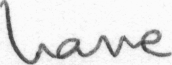} &
        \includegraphics[height=0.6cm,valign=c]{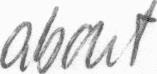}\\
        \midrule
        \checkmark & - & - & 364.10 & \includegraphics[height=0.6cm,valign=c]{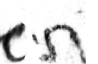} &
        \includegraphics[height=0.6cm,valign=c]{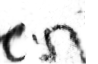} &
        \includegraphics[height=0.6cm,valign=c]{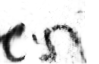} &
        \includegraphics[height=0.6cm,valign=c]{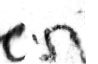}\\
        \checkmark & \checkmark & - & 207.47 &
        \includegraphics[height=0.6cm,valign=c]{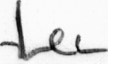} &
        \includegraphics[height=0.6cm,valign=c]{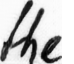} &
        \includegraphics[height=0.6cm,valign=c]{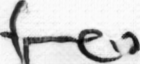} &
        \includegraphics[height=0.6cm,valign=c]{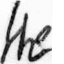}\\
        \checkmark & - & \checkmark & 138.80 &
        \includegraphics[height=0.6cm,valign=c]{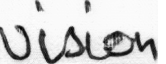} &
        \includegraphics[height=0.6cm,valign=c]{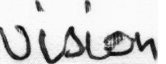} &
        \includegraphics[height=0.6cm,valign=c]{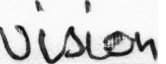} &
        \includegraphics[height=0.6cm,valign=c]{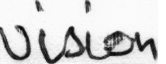}\\
        \checkmark & \checkmark & \checkmark & \textbf{130.68} &
        \includegraphics[height=0.6cm,valign=c]{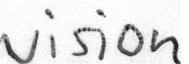} &
        \includegraphics[height=0.6cm,valign=c]{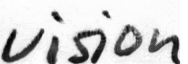} &
        \includegraphics[height=0.6cm,valign=c]{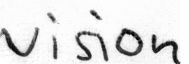} &
        \includegraphics[height=0.6cm,valign=c]{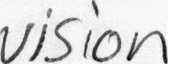}\\
        \bottomrule
    \end{tabular}
\end{table}
\endgroup

The sole use of the $\mathcal{L}_d$ leads to constantly generating an image that is able to fool the discriminator. Although the generated image looks like handwritten strokes, the content and style inputs are ignored. When combining the discriminator with the auxiliary task of writer classification $\mathcal{L}_w$, the produced results are more encouraging, but the input text is still ignored, always tending to generate the word \texttt{"the"}, since it is the most common word seen during training. When combining the discriminator with the word recognizer loss $\mathcal{L}_r$, the desired word is rendered. However, as in~\cite{alonso2019adversarial}, we suffer from the mode collapse problem, always producing unvarying word instances. When combining the three learning objectives we appreciate that we are able to correctly render the appropriate textual content while mimicking the input styles, producing diverse results. We appreciate that the FID measure also decreases for each successive combination. 

%%%%%%%%%%%%%%%%%%%%%%%%%%%%%%%%%%%%%%%%%%%%%%%%%%%%%%%%%%%%%%%%%%%%%%%%%
%%%%%%%%%%%%%%%%%%%%%%%%%%%%%%%%%%%%%%%%%%%%%%%%%%%%%%%%%%%%%%%%%%%%%%%%%
\subsection{Human Evaluation}
%Finally, besides providing qualitative results and evaluating the generative process with the FID measure, 
Finally, we also tested whether the generated images were actually indistinguishable from real ones by human judgments. We have conducted a human evaluation study as follows: we have asked $200$ human examiners to assess whether a set of images were written by a human or artificially generated. Appraisers were presented a total of sixty images, one at a time, and they had to choose if each of them was real of fake. We chose thirty real words from the IAM test partition from ten different writers. We then generated thirty artificial samples by using OOV textual contents and by randomly taking the previous writers as the sources for the calligraphic styles. Such sets were not curated, so the only filter was that the generated samples had to be correctly transcribed by the word recognizer network $R$. In total we collected $12,000$ responses. In Table~\ref{tab:human_study} we present the confusion matrix of the human assessments, with Accuracy (ACC), Precision (P), Recall (R), False Positive Rate (FPR) and False Omission Rate (FOR) values. The study revealed that our generative model was clearly perceived as plausible, since a great portion of the generated samples were deemed genuine. Only a $49.3\%$ of the images were properly identified, which shows a similar performance than a random binary classifier. Accuracies over different examiners were normally distributed. We also observe that the synthetically generated word images were judged more often as being real than correctly identified as fraudulent, with a final FPR of $55.4\%$.

\begin{table}[t!]
    \caption{Human evaluation plausibility experiment.}
    \label{tab:human_study}
    \centering
    \begin{tabular}{c @{\hskip 32pt} c}
    \begin{tabular}{c @{\hskip 12pt}c @{\hskip 8pt} c @{\hskip 8pt} c}
    \toprule
        \multirow{2}{*}{Actual} & \multicolumn{2}{c}{Predicted}\\
        \cmidrule{2-3}
         & Real & Fake&\\
         \midrule
         Genuine & $27.01$  & $22.99$& R: $ 54.1 $\\
         Generated & $27.69$ & $22.31$&  FPR: $55.4$\\

        &  P: $49.4$&  FOR: $50.8$& ACC: $49.3$\\
        \bottomrule
    \end{tabular}&
    \includegraphics[width=0.365\linewidth,valign=c]{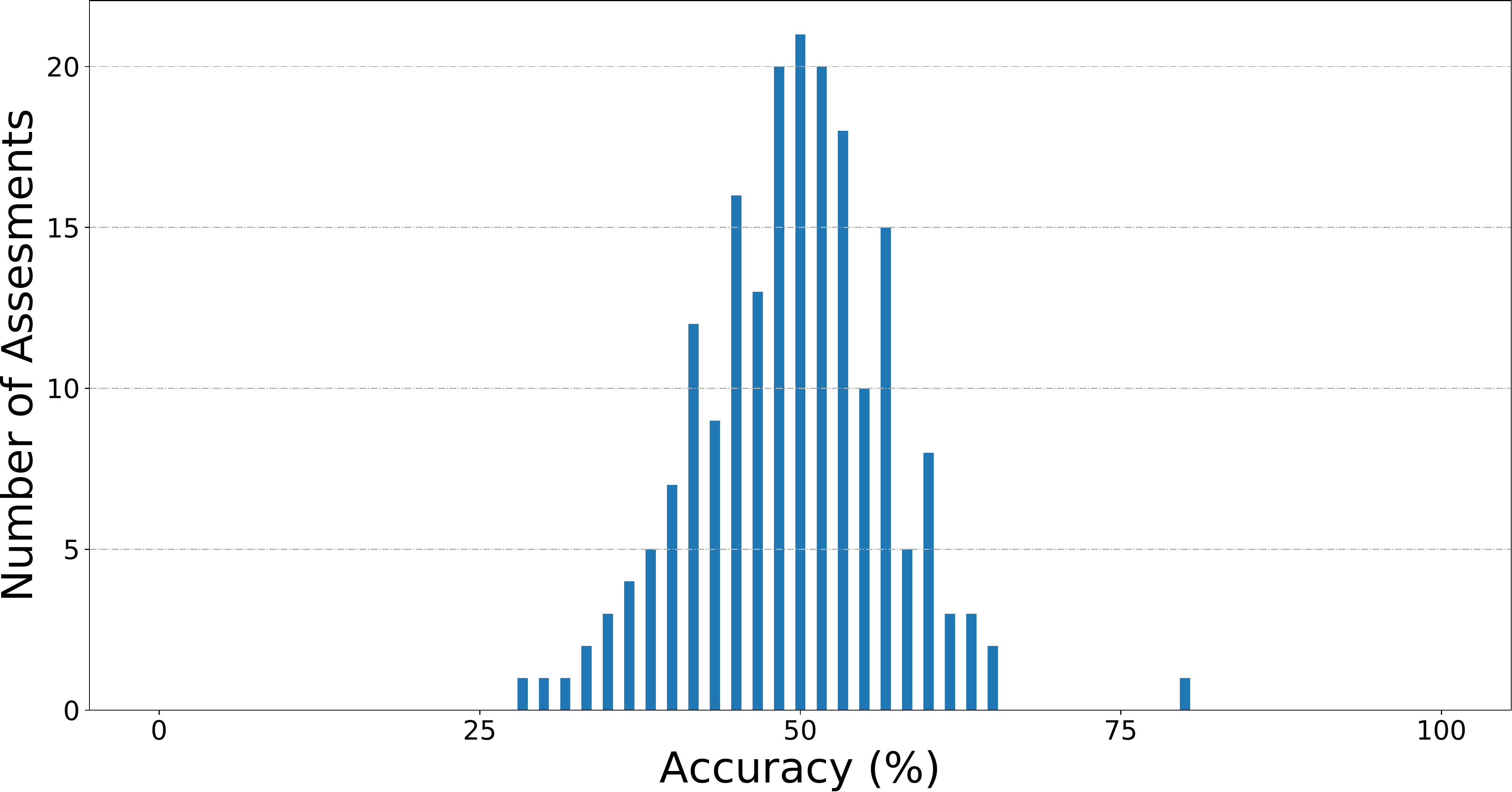}\\
    a) Confusion matrix ($\%$) &b) Accuracy distribution\\
    \end{tabular}
\end{table}

%%%%%%%%%%%%%%%%%%%%%%%%%%%%%%%%%%%%%%%%%%%%%%%%%%%%%%%%%%%%%%%%%%%%%%%%%
%%%%%%%%%%%%%%%%%%%%%%%%%%%%%%%%%%%%%%%%%%%%%%%%%%%%%%%%%%%%%%%%%%%%%%%%%
\section{Conclusion}
We have presented a novel image generation architecture that produces realistic and varied artificially rendered samples of handwritten words. Our pipeline can yield credible word images by conditioning the generative process with both calligraphic style features and textual content. Furthermore, by jointly guiding our generator with three different cues: a discriminator, a style classifier and a content recognizer, our model is able to render any input word, not depending on any predefined vocabulary, while incorporating calligraphic styles in a few-shot setup. Experimental results demonstrate that the proposed method yields images with such a great realistic quality that are indistinguishable by humans.

\section*{Acknowledgements}
This work was supported by EU H2020 SME Instrument project 849628, the Spanish projects TIN2017-89779-P and RTI2018-095645-B-C21, and grants 2016-DI-087, FPU15/06264 and RYC-2014-16831. Titan GPU was donated by NVIDIA.

%This work was partially supported by the EU H2020 SME Instrument project 849628, the grant 2016-DI-087 from the Secretaria d'Universitats i Recerca del Departament d'Economia i Coneixement de la Generalitat de Catalunya, the Spanish projects TIN2017-89779-P and RTI2018-095645-B-C21, and the grants FPU15/06264 and RYC-2014-16831. The Titan XP was donated by NVIDIA.

%\clearpage
\bibliographystyle{splncs04}
\bibliography{egbib}
\end{document}

\pagestyle{headings}
\mainmatter
\def\ECCVSubNumber{4304}  % Insert your submission number here

\title{GANwriting: Content-Conditioned Generation of Styled Handwritten Word Images\\\textemdash\\Supplementary Material}

\begin{comment}
\titlerunning{ECCV-20 submission ID \ECCVSubNumber} 
\authorrunning{ECCV-20 submission ID \ECCVSubNumber} 
\author{Anonymous ECCV submission}
\institute{Paper ID \ECCVSubNumber}
\end{comment}

\titlerunning{GANwriting: Content-Conditioned Generation of Styled Handwriting}
\author{Lei Kang$^{* \dag}$, Pau Riba$^{*}$, Yaxing Wang$^{*}$, Mar\c{c}al Rusi{\~n}ol$^{*}$, Alicia Forn\'{e}s$^{*}$, Mauricio Villegas$^{\dag}$}
% $^{*}$Computer Vision Center, Universitat Aut{\`o}noma de Barcelona, Spain\\
% {\tt\small \{lkang, priba, yaxing, marcal, afornes\}@cvc.uab.es}
% \\
% $^{\dag}$omni:us, Berlin, Germany\\
% {\tt\small \{lei, mauricio\}@omnius.com}
% }

\institute{$^{*}$Computer Vision Center, Universitat Aut{\`o}noma de Barcelona, Spain\\
{\tt\small \{lkang, priba, yaxing, marcal, afornes\}@cvc.uab.es}
\\
$^{\dag}$omni:us, Berlin, Germany\\
{\tt\small \{lei, mauricio\}@omnius.com}\\
}

\authorrunning{L. Kang et al.}

%******************
\maketitle

%%%%%%%%%%%%%%%%%%%%%%%%%%%%%%%%%%%%%%%%%%%%%%%%%%%%%%%%%%%%%%%%%%%%%%%%%
%%%%%%%%%%%%%%%%%%%%%%%%%%%%%%%%%%%%%%%%%%%%%%%%%%%%%%%%%%%%%%%%%%%%%%%%%
\section{Video Interpolation}
\begin{figure}[ht!]
    \centering
    \begin{tabular}{c}
        \frame{\includegraphics[width=0.9\linewidth,valign=c]{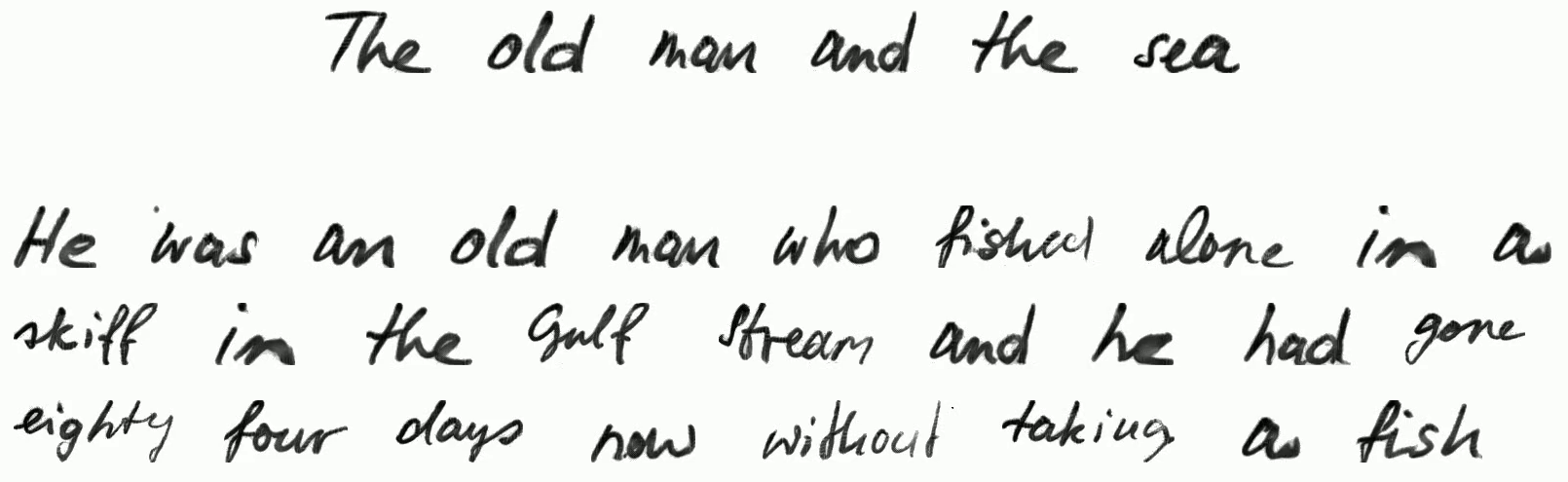}}\\
        \frame{\includegraphics[width=0.9\linewidth,valign=c]{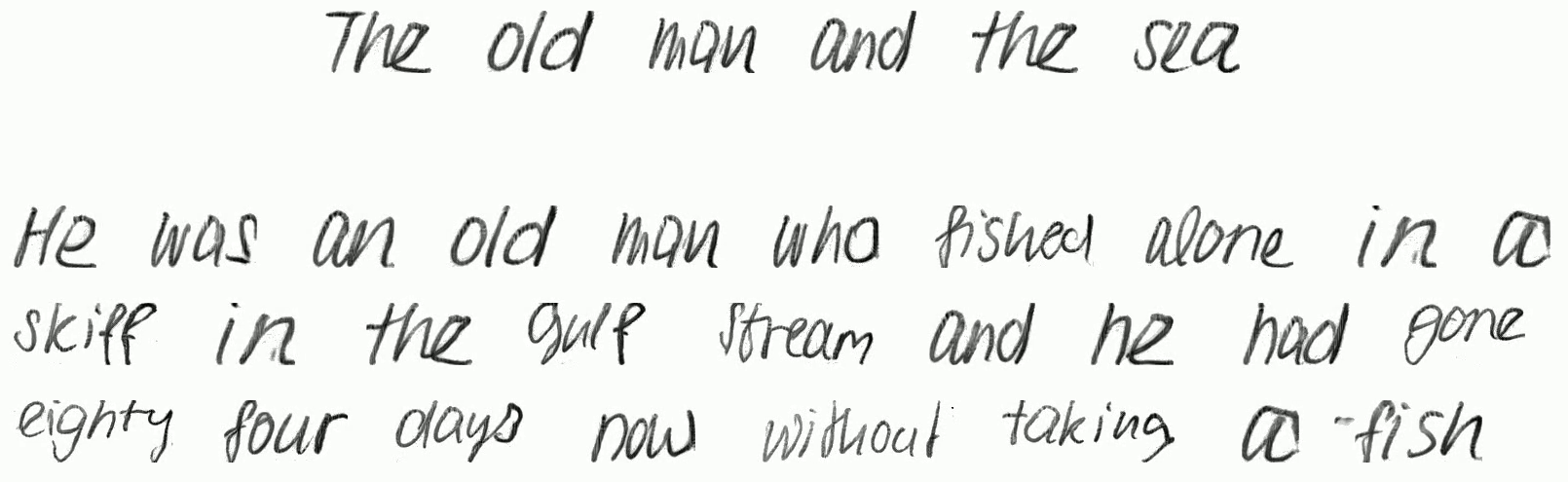}}\\
 \end{tabular}
\caption{Sample frames of the interpolation video.}
    \label{tab:screenshots}
\end{figure}

To better showcase the meaningfulness of the learned stylistic embedding space, find attached a video where we animate a much finer interpolation than the one pictured in the paper, between different calligraphic styles of several words, composing the first sentence of Ernest Hemingway's \emph{``The Old Man and The Sea''}. We appreciate how the generator is able to provide a smooth transition between different writing styles for a given static content. We provide some screenshots of such video in Figure~\ref{tab:screenshots}.

%%%%%%%%%%%%%%%%%%%%%%%%%%%%%%%%%%%%%%%%%%%%%%%%%%%%%%%%%%%%%%%%%%%%%%%%%
%%%%%%%%%%%%%%%%%%%%%%%%%%%%%%%%%%%%%%%%%%%%%%%%%%%%%%%%%%%%%%%%%%%%%%%%%
\section{Limitation of the proposed method when dealing with calligraphic styles}

\begin{figure}[ht!]
    \centering
    \resizebox{\textwidth}{!}{
    \begin{tabular}{c@{\hskip 6pt}ccc}
    \toprule
    \multirow{4}{*}{\includegraphics[width=0.42\linewidth,valign=c]{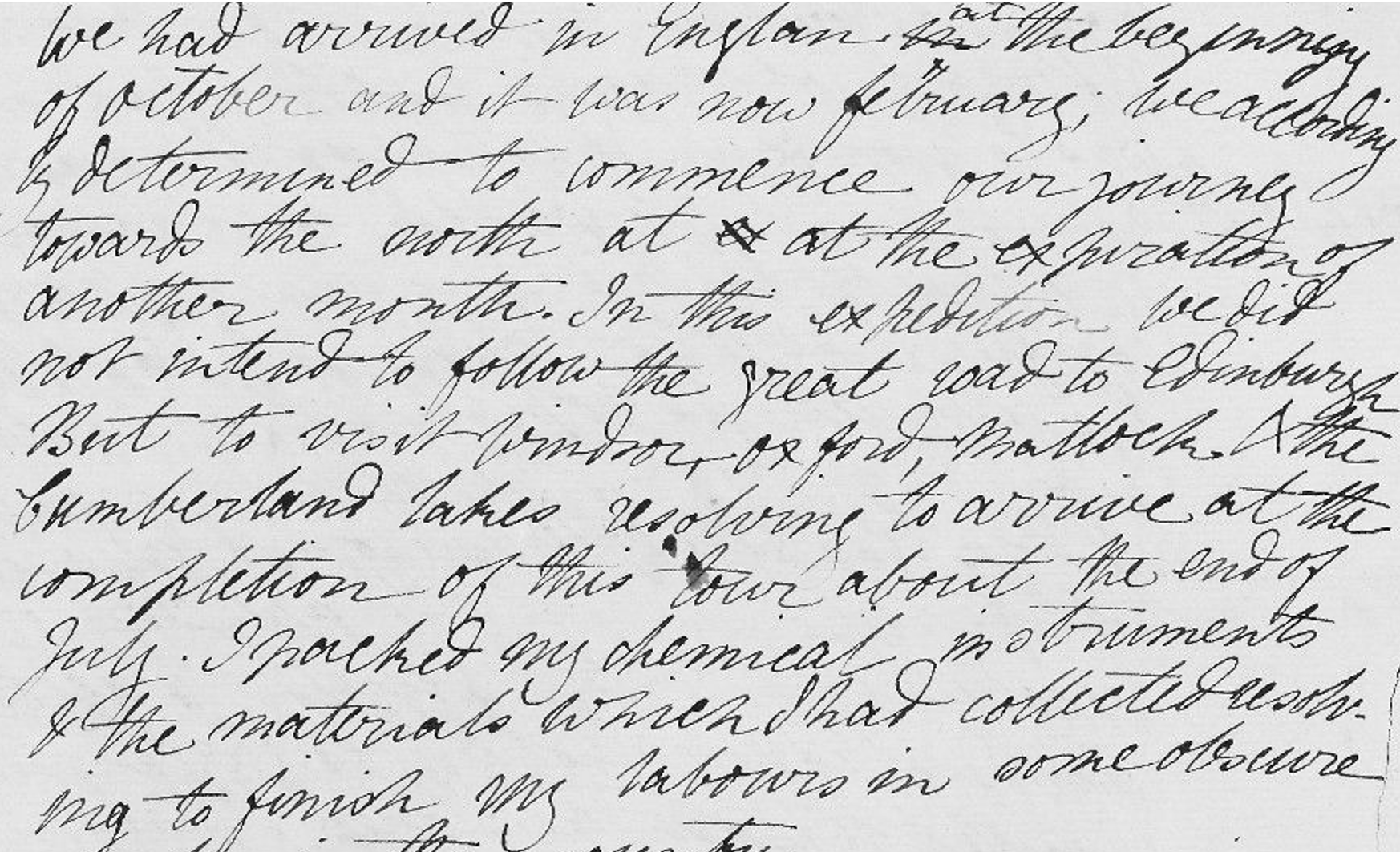}}&
     \includegraphics[width=0.18\linewidth,valign=c]{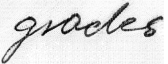}&
     \includegraphics[width=0.18\linewidth,valign=c]{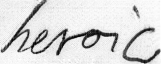}&
     \includegraphics[width=0.18\linewidth,valign=c]{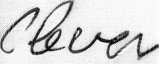} \\
     &\includegraphics[width=0.18\linewidth,valign=c]{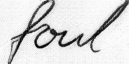}&
     \includegraphics[width=0.18\linewidth,valign=c]{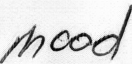}&
     \includegraphics[width=0.18\linewidth,valign=c]{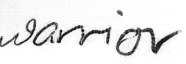} \\
     &\includegraphics[width=0.18\linewidth,valign=c]{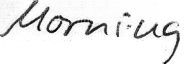}&
     \includegraphics[width=0.18\linewidth,valign=c]{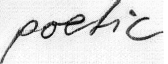}&
     \includegraphics[width=0.18\linewidth,valign=c]{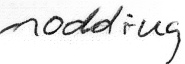}\\
     &\includegraphics[width=0.18\linewidth,valign=c]{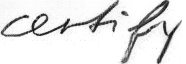}&
     \includegraphics[width=0.18\linewidth,valign=c]{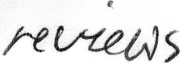}&
     \includegraphics[width=0.18\linewidth,valign=c]{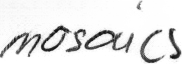}\\
         Original manuscript & \multicolumn{3}{c}{Generated samples}\\
     \bottomrule
    \end{tabular}
    }
\caption{Limitations of the proposed approach when mimicking Mary Shelley's handwriting style.}
    \label{tab:limits}
\end{figure}

We evidence in Figure~\ref{tab:limits} the limitations of the proposed approach on imitating calligraphic styles. Unlike in~\cite{haines2016my}, where characteristic glyphs from a given writer were manually cropped to perfectly compose a fraudulent text excerpt as if it was written by a certain person, our approach is not able to produce such levels of mimicking. When the model, trained with the IAM dataset, is fed with an unconventional calligraphic style, the proposed approach is not able to convey such stylistic aspects to the generated word samples. In Figure~\ref{tab:limits}, we injected word samples written by Mary Shelley, and, the reader will appreciate how the rendered results are not able to imitate the visual aspect of such handwriting. However, the proposed generative method is still able to correctly render the textual contents, regardless of the provided calligraphic style.

%%%%%%%%%%%%%%%%%%%%%%%%%%%%%%%%%%%%%%%%%%%%%%%%%%%%%%%%%%%%%%%%%%%%%%%%%
%%%%%%%%%%%%%%%%%%%%%%%%%%%%%%%%%%%%%%%%%%%%%%%%%%%%%%%%%%%%%%%%%%%%%%%%%
\section{Qualitative comparison with Alonso \emph{et al.}~\cite{alonso2019adversarial}}

We present in Table~\ref{tab:alonso}, a qualitative comparison with the work of Alonso \emph{et al.}~\cite{alonso2019adversarial}. We can appreciate how our proposed method clearly produces much credible generated images while being able to render the same content word with different calligraphic styles. Whereas~\cite{alonso2019adversarial} suffers from the mode collapse problem, always tending towards producing similar glyphs, our proposed method is able to yield different stylistic instances of the same textual content.

\newlength{\myheight}
\setlength{\myheight}{5ex}
\begin{table}[ht!]
\caption{Qualitative comparison with Alonso \emph{et al.}. Images reprinted from~\cite{alonso2019adversarial}.}
    \label{tab:alonso}
    \centering
    \begin{tabular}{c@{\hskip 8pt}c cc c c@{\hskip 8pt} c}
    \toprule
     \multirow{2}{*}{Content} && \multirow{2}{*}{Alonso \emph{et al.}~\cite{alonso2019adversarial}} & & \multicolumn{3}{c}{Ours}\\
     \cmidrule{5-7}
     &&&& Style $A$ & Style $B$ & Style $C$\\
     \cmidrule[\lightrulewidth]{1-1} \cmidrule[\lightrulewidth]{3-3} \cmidrule[\lightrulewidth]{5-7}
        \texttt{"olibus"}&&
        \includegraphics[height=5ex,valign=c]{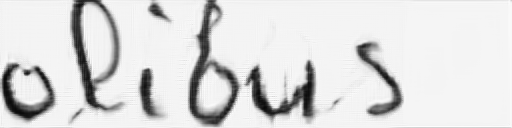} & &
        \includegraphics[height=\myheight,valign=c]{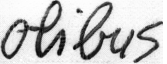} &
        \includegraphics[height=\myheight,valign=c]{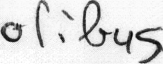} & 
        \includegraphics[height=\myheight,valign=c]{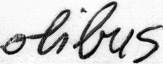} \\ 
        \midrule
        \texttt{"reparer"}&&
        \includegraphics[height=\myheight,valign=c]{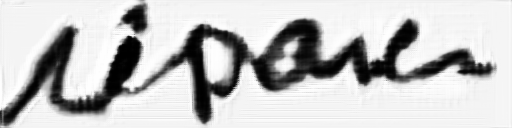} & &
        \includegraphics[height=\myheight,valign=c]{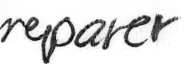} &  
        \includegraphics[height=\myheight,valign=c]{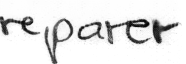} &
        \includegraphics[height=\myheight,valign=c]{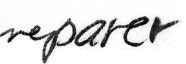} \\ 
        \midrule
        \texttt{"bonjour"}&&
        \includegraphics[height=\myheight,valign=c]{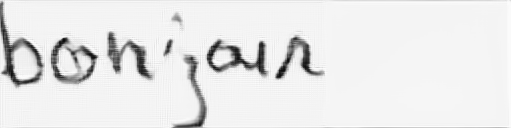} & &
        \includegraphics[height=\myheight,valign=c]{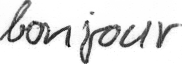} &  
        \includegraphics[height=\myheight,valign=c]{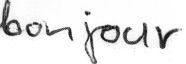} &
        \includegraphics[height=\myheight,valign=c]{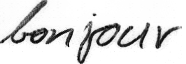} \\  
        \midrule
        \texttt{"famille"}&&
        \includegraphics[height=\myheight,valign=c]{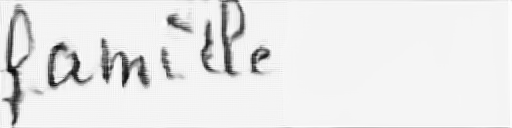} & &
        \includegraphics[height=\myheight,valign=c]{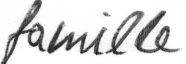} &  
        \includegraphics[height=\myheight,valign=c]{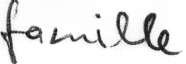} & 
        \includegraphics[height=\myheight,valign=c]{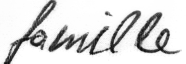} \\
        \midrule
        \texttt{"gorille"}&&
        \includegraphics[height=\myheight,valign=c]{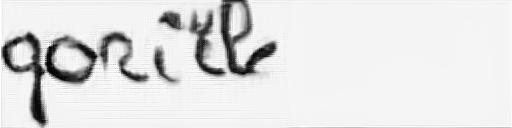} & &
        \includegraphics[height=\myheight,valign=c]{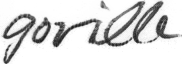} &  
        \includegraphics[height=\myheight,valign=c]{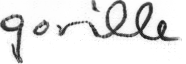} & 
        \includegraphics[height=\myheight,valign=c]{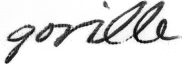} \\  
        \midrule
        \texttt{"malade"}&&
        \includegraphics[height=\myheight,valign=c]{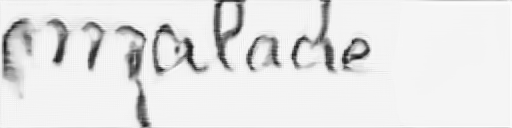} & &
        \includegraphics[height=\myheight,valign=c]{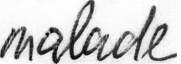} &  
        \includegraphics[height=\myheight,valign=c]{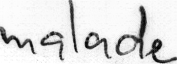} & 
        \includegraphics[height=\myheight,valign=c]{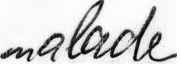} \\  
        \midrule
        \texttt{"certes"}&&
        \includegraphics[height=\myheight,valign=c]{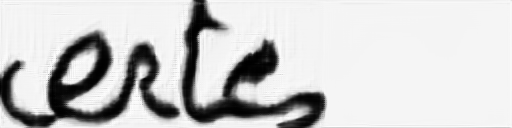} & &
        \includegraphics[height=\myheight,valign=c]{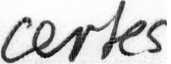} &  
        \includegraphics[height=\myheight,valign=c]{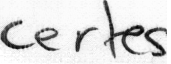} & 
        \includegraphics[height=\myheight,valign=c]{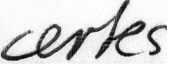} \\ 
        \midrule
        \texttt{"golf"}&&
        \includegraphics[height=\myheight,valign=c]{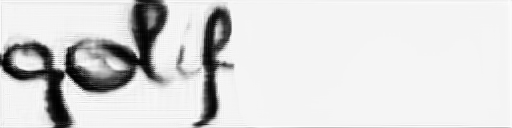} & &
        \includegraphics[height=\myheight,valign=c]{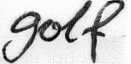} &  
        \includegraphics[height=\myheight,valign=c]{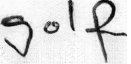} & 
        \includegraphics[height=\myheight,valign=c]{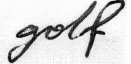} \\ 
        \midrule
        \texttt{"des"}&&
        \includegraphics[height=\myheight,valign=c]{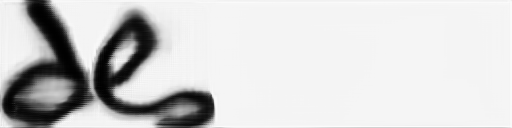} & &
        \includegraphics[height=\myheight,valign=c]{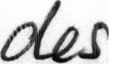} &  
        \includegraphics[height=\myheight,valign=c]{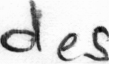} & 
        \includegraphics[height=\myheight,valign=c]{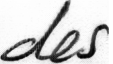} \\  
        \midrule
        \texttt{"ski"}&&
        \includegraphics[height=\myheight,valign=c]{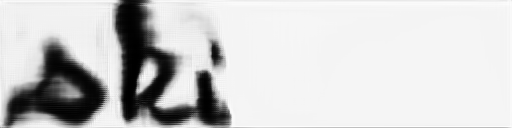} & &
        \includegraphics[height=\myheight,valign=c]{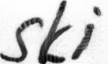} &  
        \includegraphics[height=\myheight,valign=c]{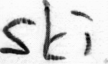} & 
        \includegraphics[height=\myheight,valign=c]{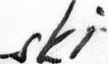} \\  
        \midrule
        \texttt{"le"}&&
        \includegraphics[height=\myheight,valign=c]{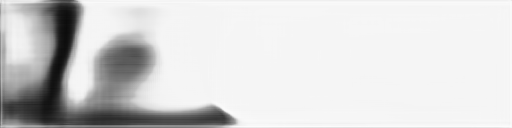} &  &
        \includegraphics[height=\myheight,valign=c]{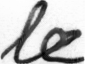} &  
        \includegraphics[height=\myheight,valign=c]{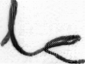} &
        \includegraphics[height=\myheight,valign=c]{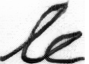} \\  
        \bottomrule
\end{tabular}
\end{table}

%%%%%%%%%%%%%%%%%%%%%%%%%%%%%%%%%%%%%%%%%%%%%%%%%%%%%%%%%%%%%%%%%%%%%%%%%
%%%%%%%%%%%%%%%%%%%%%%%%%%%%%%%%%%%%%%%%%%%%%%%%%%%%%%%%%%%%%%%%%%%%%%%%%
\section{t-SNE Embedding visualizations}

\begin{figure}[ht!]
    \centering
    \includegraphics[angle=90,height=.93\textheight]{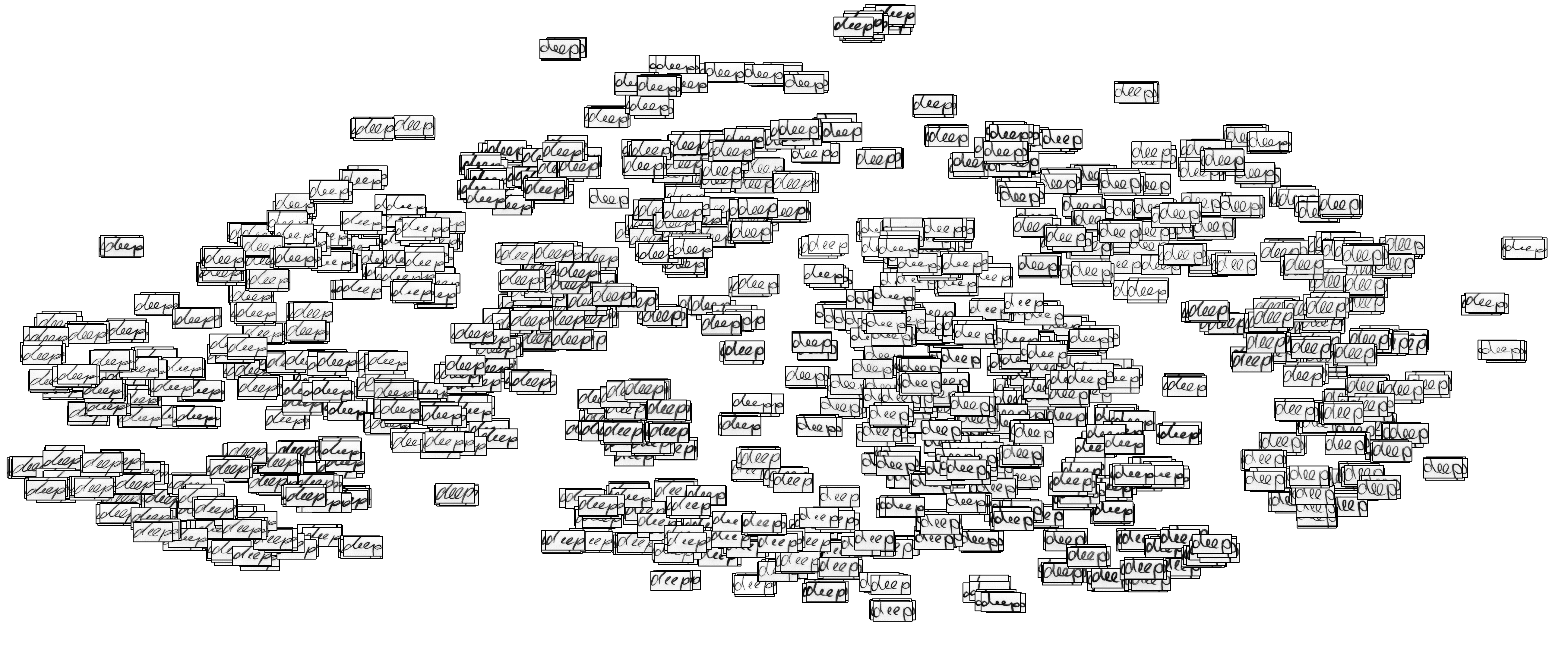}
    \caption{t-SNE embedding visualization of $2.500$ generated instances of the word \texttt{"deep"}.}
    \label{fig:tsne0}
\end{figure}

\begin{figure}[ht!]
    \centering
    \includegraphics[angle=90,height=.93\textheight]{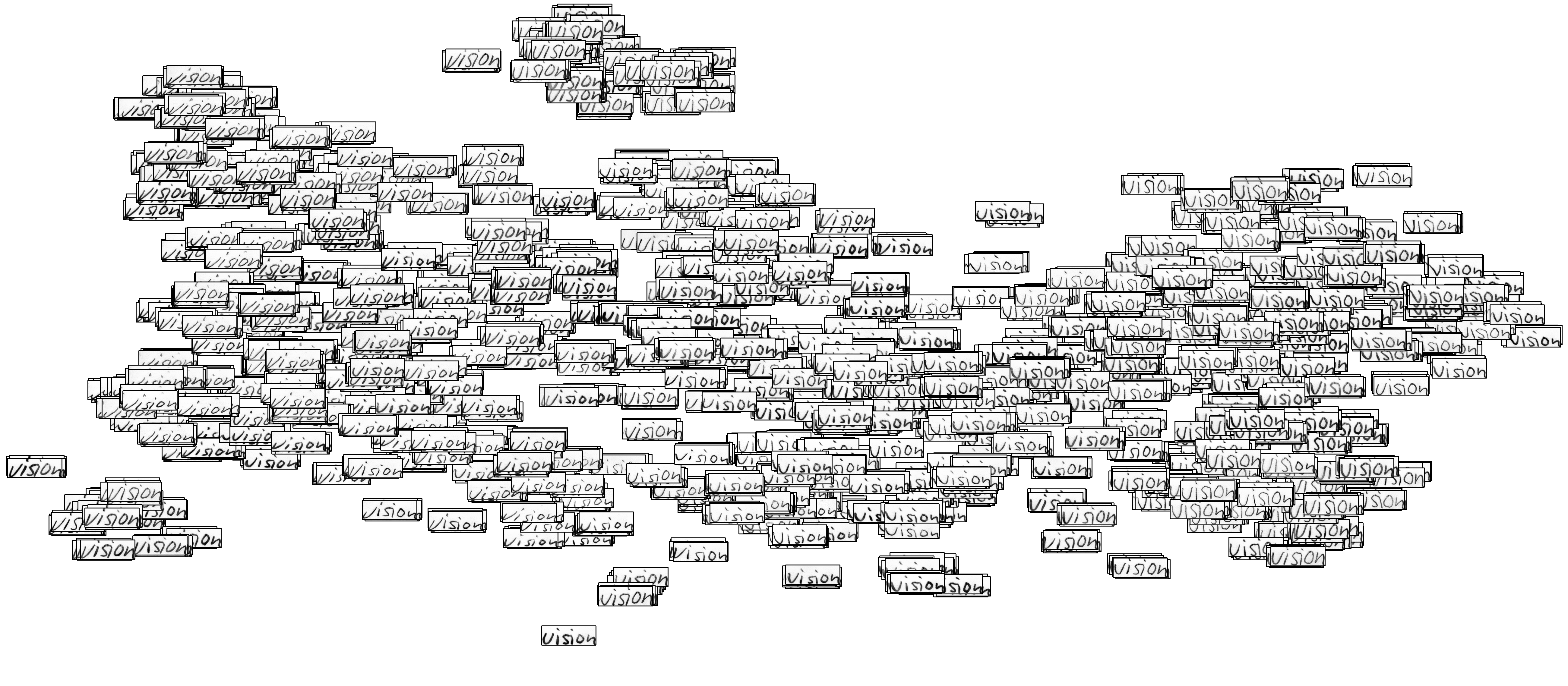}
    \caption{t-SNE embedding visualization of $2.500$ generated instances of the word \texttt{"vision"}.}
    \label{fig:tsne1}
\end{figure}

\begin{figure}[ht!]
    \centering
    \includegraphics[angle=90,height=.93\textheight]{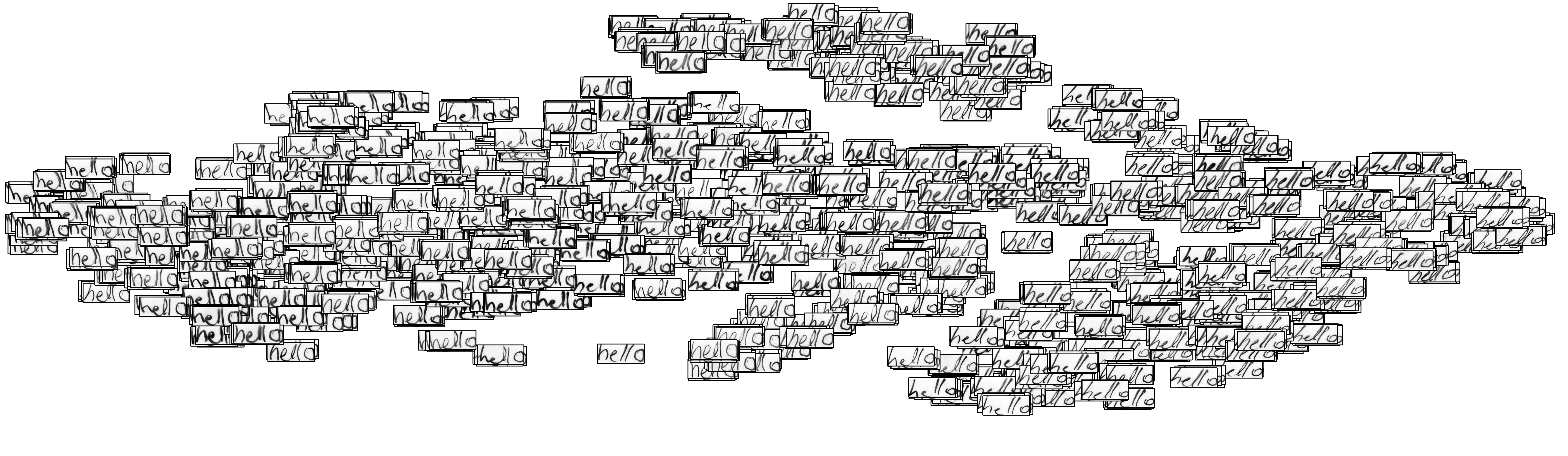}
    \caption{t-SNE embedding visualization of $2.500$ generated instances of the word \texttt{"hello"}.}
    \label{fig:tsne2}
\end{figure}

\begin{figure}[ht!]
    \centering
    \includegraphics[angle=90,height=.93\textheight]{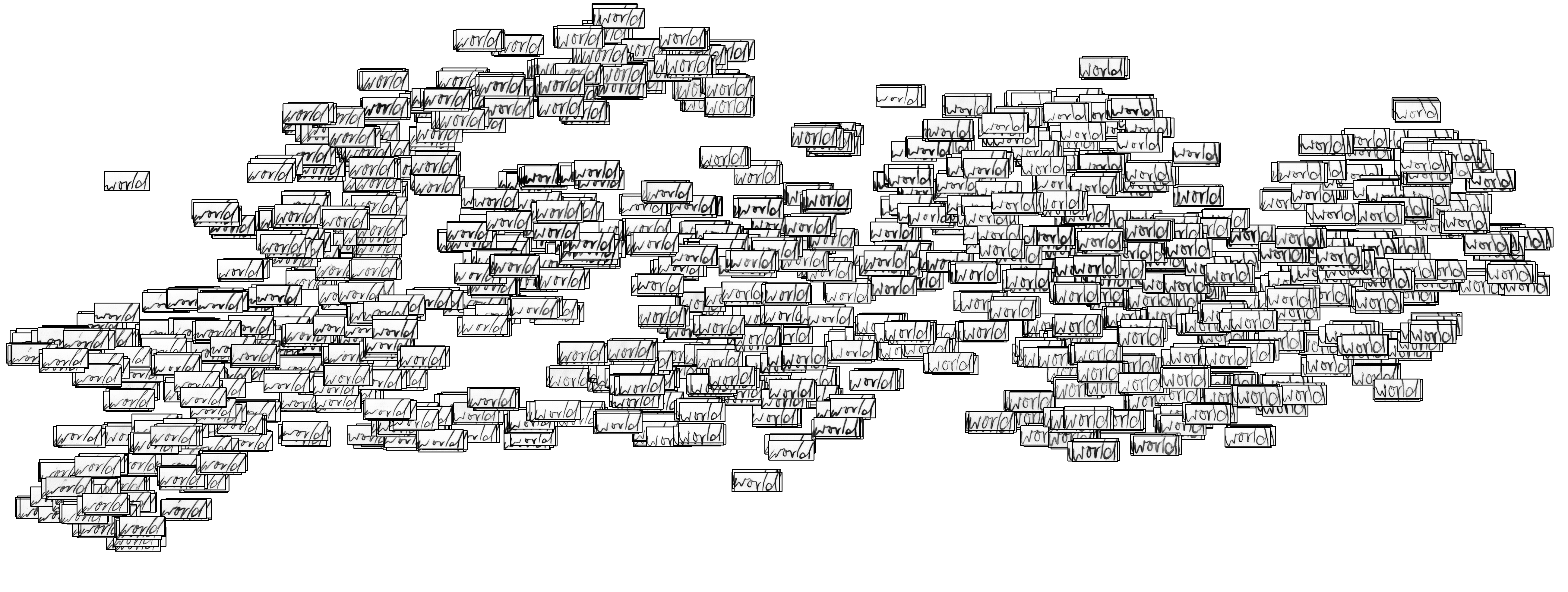}
    \caption{t-SNE embedding visualization of $2.500$ generated instances of the word \texttt{"world"}.}
    \label{fig:tsne3}
\end{figure}

 Due to space constrains, we are aware that the t-SNE plot presented in the paper in Figure 5 is shown at a quite small scale. This difficult its inspection. We provide here in Figures~\ref{fig:tsne0},~\ref{fig:tsne1},~\ref{fig:tsne2} and~\ref{fig:tsne3}, four different t-SNE plots for images generated with the same textual content and for various calligraphic styles.

\bibliographystyle{splncs04}
\bibliography{egbibsup}